\documentclass[times, review, 10pt]{elsarticle}
\usepackage{amsmath,amsfonts, amssymb}
\usepackage{algorithmic}
\usepackage{array}
\usepackage[caption=false,font=normalsize,labelfont=sf,textfont=sf]{subfig}
\usepackage{textcomp}
\usepackage{stfloats}
\usepackage{url}
\usepackage{booktabs}
\usepackage{verbatim}
\usepackage{graphicx}
\usepackage{adjustbox}
\usepackage{booktabs}
\usepackage{multirow}
\usepackage{caption}
\usepackage{xcolor}
\usepackage{natbib}
\usepackage[normalem]{ulem}

\usepackage{amssymb}
\usepackage{amsmath}

\journal{Pattern Recognition}

\begin{document}

\begin{frontmatter}



\title{Water Reflection Detection Using Symmetric Attention}

\author[label1]{Shuxuan Yao}
\ead{yaoshuxuan@stu.ouc.edu.cn}

\author[label2]{Chengjia Wang}
\ead{chengjia.wang@hw.ac.uk}

\author[label3]{Jianyuan Sun}
\ead{jianyuan.sun@dmu.ac.uk}

\author[label1]{Junyu Dong}
\ead{dongjunyu@ouc.edu.cn}

\author[label1]{Xinghui Dong\corref{cor1}}
\ead{xinghui.dong@ouc.edu.cn}

\cortext[cor1]{Corresponding author}

\affiliation[label1]{
  organization={State Key Laboratory of Physical Oceanography and the Faculty of Information Science and Engineering, Ocean University of China},
  addressline={238 Songling Road},
  city={Qingdao},
  postcode={266100},
  state={Shandong},
  country={China}
}

\affiliation[label2]{organization={School of Mathematical and Computer Sciences, Heriot-Watt University},
             addressline={Riccarton Mains Road, EH14 4AS, Edinburgh},
             country={United Kingdom}}

\affiliation[label3]{organization={School of Computer Science and Informatics, De Montfort University},
             addressline={Leicester, LE1 9BH},
             country={United Kingdom}}

\tnotetext[Github]{Code and models will be made available at https://github.com/INDTLab/SAWRD-Net.}

\begin{abstract}
Reflections of water pose a significant challenge for computer vision systems, as standard deep learning models frequently confuse objects with their mirror images, producing spurious false positives and negatives in tasks such as object detection and semantic segmentation. As a result, detecting reflection axes in natural-water scenes is pivotal for reliable object detection and scene understanding. To mitigate this issue, we leverage the intrinsic imperfect reflective symmetry of water and introduce a Symmetry-Aware Water Reflection Detection Network, namely, SAWRD-Net, that couples dihedral group–equivariant convolutions with a matrix-decomposition decoder in an end-to-end framework. First, dihedral group convolutional layers extract geometry-consistent feature maps that explicitly encode both rotational and mirror symmetries. A Multi-scale Reflection Equivariant block then aggregates features across scales and employs a symmetric-attention mechanism to highlight reflection-relevant regions. The proposed matrix-decomposition decoder factorizes high-dimensional features into compact low-rank parameter and confidence spaces, after which the network directly regresses keypoints on the reflection axis. Then a robust principal component analysis fits the final axis. Evaluated on the largest available water reflection scene data set, SAWRD-Net achieves a true-positive rate of 0.890 against human annotations, outperforming all existing water reflection detectors.
\end{abstract}


\begin{highlights}
\item We introduce an end-to-end water reflection detection network, i.e., SAWRD-Net, which couples group-equivariant convolutions with low-rank matrix decomposition 
\item We develop a symmetric-attention mechanism that redistributes feature weights according to reflection cues, thereby improving the capture of symmetry-specific information and enhancing detection accuracy
\item We design an MSRE block and an MD Decoder, which leverages the intrinsic symmetry of water reflection scenes while modeling long-range context via precise low-rank factorization
\end{highlights}


\end{frontmatter}



\section{Introduction}
\label{Introduction}
Water reflections are ubiquitous in nature photography and constitute a prominent class of outdoor images \cite{dong2024wrd,zhong2013water,zhang2010water,zhong2011water}, thus accurately modeling them benefits large-scale visual understanding \cite{nguyen2022reflection}. Reflections of water create mirror-like duplicates of scene objects \cite{zha2024weakly,huang2023symmetry}. Ideally, the object and its reflection would exhibit perfect reflective symmetry on the water surface, but surface ripples, refraction, and light scattering introduce \emph{imperfect (or approximate) symmetry} cause an object and its virtual image to exhibit a geometric correspondence relative to the water surface \cite{zhang2010water}, forming a reflective symmetry \cite{podgorelec2024ahils}. Detecting the underlying reflection axis and separating real objects from their virtual counterparts is therefore a prerequisite for robust scene parsing, object detection \cite{tang2022learning,tang2024divide}, and image segmentation in natural environments \cite{dong2024wrd}.



\begin{figure}[ht]
  \centering
  \includegraphics[width=1.0\linewidth,keepaspectratio]{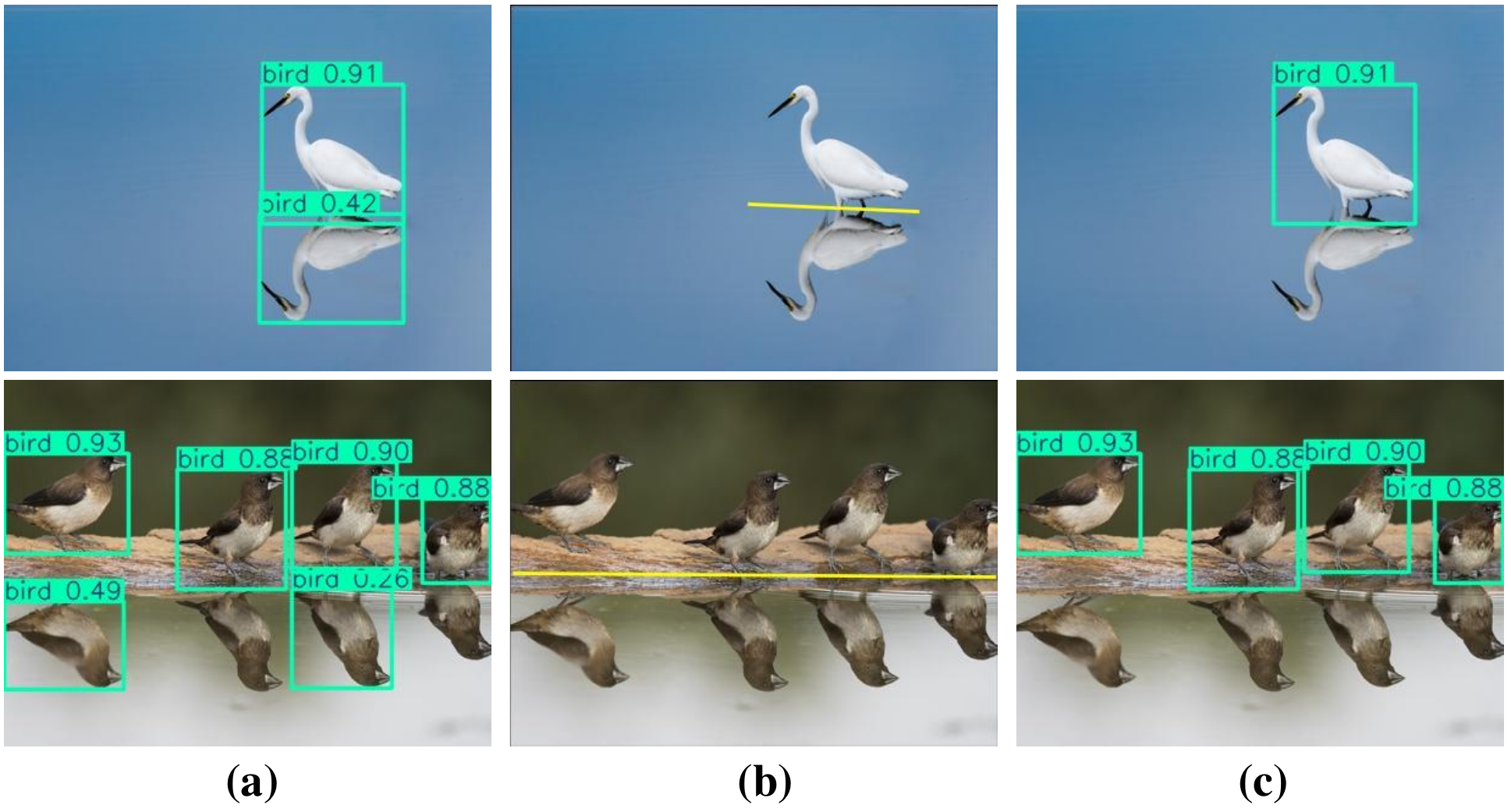}  
  \caption{The case of failure in detecting water reflection images using the YOLOv12 \cite{tian2025yolov12} object detector (a), the reflection axis detected by the SAWRD-Net (b), and the successful detection result of the YOLOv12 \cite{tian2025yolov12} object detector after filtering the reflection area of the water reflection image based on the reflection axis detected by the SAWRD-Net (c).}
  \label{fig:OR}
\end{figure}

Water reflection images exhibit mirror symmetry, causing segmentation and detection networks to confuse real objects with their reflections \cite{huang2022vision}. As shown in Fig.~\ref{fig:OR}, YOLOv12 \cite{tian2025yolov12} identifies an object and its reflection as separate instances, thus generating false positives. By localizing the reflection axis and suppressing the mirrored region, such errors can be eliminated, enabling correct recognition. In this situation, a dedicated preprocessing is necessary to identify and filter out reflective regions to achieve reliable scene understanding.

Water reflection detection originally relied on manual feature-based pipelines \cite{zhong2013water,zhang2010water,zhong2011water}. These approaches require expert tuning of gradient thresholds, color ratios, and geometric heuristics, which sharply limits their applicability in real-world scenarios. Deep learning models, in contrast, learn discriminative features directly from data and therefore offer greater robustness. However, comprehensive benchmarks and large-scale data sets for this purpose remain scarce. In the literature, the only end-to-end water reflection detection model is WRD-Net \cite{dong2024wrd}, which uses parallel attention for multi-scale context modeling, but implicitly learns symmetry instead of encoding mirror geometry. On the other hand, EquiSym \cite{seo2022reflection} and related detectors \cite{funk2017beyond,seo2021learning} enforce strict equivariance under perfect symmetry, making them vulnerable to water ripples and refraction. In contrast, SAWRD-Net balances reflection equivariance and perturbation tolerance via equivariant convolutions and low-rank decomposition.

Earlier vision pipelines achieved rotational or mirror equivariance by pairing orientation-sensitive key-point detectors with an orientation-normalization step, for example, aligning feature patches to their dominant gradient direction \cite{loy2006detecting,prasad2005detecting}. The goal was to preserve one-to-one correspondences between features when the input image was rotated or flipped. Although effective for shallow, gradient-based descriptors, this mechanism degrades when transferred to modern deep networks. Deep feature maps undergo complex, non-linear distortions under the same transformations, breaking the assumed correspondence. Equivariant learning addresses this limitation by embedding the transformation law directly into the network. By guaranteeing predictable, structurally consistent changes throughout all layers whenever the input is rotated or reflected, it provides a principled foundation for symmetry-aware tasks and naturally respects the geometric constraints of water reflection scenes, especially in high-resolution outdoor footage captured under uncontrolled lighting \cite{cohen2016group}.



Water reflection detection focuses on modeling global yet imperfect symmetry in natural scenes. Such symmetry, caused by ripples, refraction, etc., implies object and reflection are not pixel-wise identical but globally correlated. Since matrix decomposition uses low-rank constraints to capture global symmetry while treating local perturbations as residuals, it naturally formulates approximate symmetry in water reflection scenes. Building on the established applicability of group-equivariant convolutions to symmetry detection, we construct a Multi-scale Reflection Equivariant (MSRE) block that extracts rotation- and reflection-equivariant features, complemented by a symmetric-attention mechanism that concentrates on reflection-relevant regions. Since water reflections form a global level structure and typically exhibit incomplete reflective symmetry, we further introduce a dedicated decoder, the Matrix Decomposition (MD) Decoder, which factorises the high-dimensional representation into complementary low-rank parameter and confidence sub-spaces, thereby enabling more reliable axis estimation in challenging, real-world water reflection scenes.

The main contributions of this work are threefold. 
\begin{enumerate}
    \item We introduce SAWRD-Net, an end-to-end network for water reflection detection that couples group-equivariant convolutions with low-rank matrix decomposition. This network regresses symmetry-axis point sets and refines the axis through principal component analysis.
    \item We develop a symmetric-attention mechanism that redistributes feature weights according to reflection cues, thereby improving the capture of symmetry-specific information and enhancing detection accuracy.
    \item We design an MSRE block together with an MD Decoder, which leverages the intrinsic symmetry of water scenes while modeling long-range context via precise low-rank factorization, yielding robust axis estimation under imperfect symmetry.
\end{enumerate}

The remainder of the paper is organized as follows:
Section 2 reviews related work. Section 3 describes the proposed SAWRD-Net and its symmetric-attention module. Section 4 details the experimental setup, and Section 5 reports the results. Section 6 concludes the paper and provides a further detailed discussion.

\section{Related Work}
\subsection{Water Reflection Detection}
\label{Water Reflection Detection}
Water reflection detection techniques have progressed from manual feature engineering to modern deep learning approaches. Early work by \citet{zhang2010water} introduced a \emph{flip-invariant shape descriptor} to mitigate distortions caused by surface waves. In separate work, \citet{zhong2011water} proposed a motion blur invariant feature space (MBIS) with a dynamic programming axis detector; they later combined MBIS with curvelet coefficients in a dual-channel algorithm \citet{zhong2013water}. These methods exploit log–polar representations and dynamic programming but remain limited by hand-crafted features. More recently, \citet{dong2024wrd} framed water reflection detection as a symmetry axis point prediction task and proposed WRD-Net, a deep architecture that fuses a parallel-attention Vision Transformer with an \emph{atrous} spatial pyramid. Although WRD-Net removes the need for manual features, it under-utilizes the inherent reflective symmetry of water scenes.


\subsection{Reflection Symmetry Detection}
\label{Reflection Symmetry Detection}
Traditional methods usually generate initial symmetry assumptions based on geometric features, and then make iterative corrections by optimizing the symmetry coefficients or matching feature pairs. In contrast, 
Deep learning methods use neural networks to automatically learn symmetry features through end-to-end training and optimize the model parameters with the help of back-propagation to progressively improve detection accuracy.

\subsubsection{Traditional Methods}
\label{Traditional Methods}
Early methods for symmetry detection often relied on global image properties and geometric transforms. For instance, \citet{marola1989detection} proposed an algorithm that detects the reflection axis by determining the image's centre of mass and maximising symmetry coefficients. A Hough transform-based method for reflection symmetry axis detection was proposed in \cite{ogawa1991symmetry}, which converts the problem into a line detection task in Hough space, allowing for robust detection in complex line drawings. Other approaches leveraged the frequency domain. \citet{sun1999fast} proposed a method based on histograms of gradient directions and Fourier transform, while the method in \cite{keller2004algebraic} operates by examining specific patterns in an object's Fourier transform using the angular difference function (ADF). Expanding on transform techniques, an efficient geometry-based framework was proposed in \cite{tuytelaars2003noncombinatorial} that uses invariant hashing and the Hough transform to detect various symmetric configurations, including periodicity and mirror symmetry, while maintaining robustness to perspective distortions.


A second major family of traditional methods is based on local features and point matching. \citet{loy2006detecting} proposed a symmetry detection method based on this principle, where feature points are generated and matched to find symmetric pairs, detecting both mirror and rotational symmetry. This concept was extended by \citet{lee2011curved} with a local feature-based algorithm for detecting curved glide-reflection symmetry in real images. An adaptive algorithm, also based on local feature point matching, was proposed in \cite{cai2014adaptive} to detect both reflectional and rotational symmetries. Similarly, a bilateral symmetry detection method based on the discovery of extremum curvature points along the edges was proposed in \cite{atadjanov2015bilateral}, which uses histograms of curvature responses to describe the characteristics. Other advanced approaches used different strategies. \citet{elawady2017wavelet} proposed a method for the detection of high-precision global symmetry axis using the Log-Gabor wavelet transform, which integrates edge, texture and color descriptors within a voting framework. Meanwhile, \citet{cicconet2017finding} introduced a new framework called Mirror Symmetry via Registration (MSR) to detect mirror symmetry in data. Finally, a method combining appearance and gradient information was proposed in \cite{gnutti2021combining}, and its effectiveness was confirmed through user perception validation experiments.

\subsubsection{Learning-Based Methods}
Deep learning-based approaches automatically learn symmetry features through neural networks, enabling them to capture symmetry patterns at different scales and orientations with more precision. \citet{funk2017beyond} proposed a deep symmetry detector, Sym-NET, which mimics human perception of non-planar symmetry by converting sparse symmetry labels into dense heat maps for the network to predict. \citet{seo2021learning} proposed Polar Matching Convolution (PMC), which utilizes high-dimensional matching kernels and relational descriptors to identify symmetry patterns in challenging real-world images. Most relevant to our work, \citet{seo2022reflection} introduced an equivariant learning-based method. It employs an equivariant convolutional neural network to simultaneously detect reflective and rotational symmetries while maintaining detection accuracy under geometric transformations.



Previous methods have primarily focused on identifying symmetric patterns within the candidate directional space to better exploit the symmetrical characteristics in the images. To our knowledge, however, very few techniques employ deep learning for water reflection detection. Although WRD-Net \cite{dong2024wrd} was built on top of deep learning techniques, it combined parallel attention transformers with dilated spatial pyramids, thus failing to fully capitalize on the inherent reflection symmetry characteristics of water surfaces. Some symmetry detection methods, such as EquiSym \cite{seo2022reflection}, leveraged the concept of equivariant learning to construct a fully equivariant network. However, these networks were designed for the detection of perfect symmetry, whereas water reflection is a quintessential example of an imperfect reflection phenomenon. Starting from symmetry detection and designed to handle both inherent symmetry in water reflection images and noise induced by external factors like water waves, we propose SAWRD-Net to boost detection performance.

\section{Symmetry-Aware Water Reflection Detection Network}
Feature equivariance plays a fundamental role in the detection of image symmetry. Since the symmetric properties exhibited in images remain invariant under Euclidean transformations, images subjected to such transformations should maintain symmetry detection results consistent with the original images. Using this equivariance property and combining it with the inherent reflective symmetry characteristics of water reflection images \cite{weyl2015symmetry}, the equivariance to reflection emerges as a key discriminative characteristic for water reflection detection models. Considering the decisive impact of global contextual dependencies in images on water reflection detection and the nature of water reflection as an imperfect reflective symmetry, we further employ matrix decomposition \cite{geng2021attention} to extract global information from network features. This process involves constructing a lookup dictionary and its corresponding encoding while eliminating the influence of noise, thereby effectively capturing the intrinsic correlations among features.

\begin{figure}[t]
  \centering
  \includegraphics[width=0.9\linewidth,keepaspectratio]{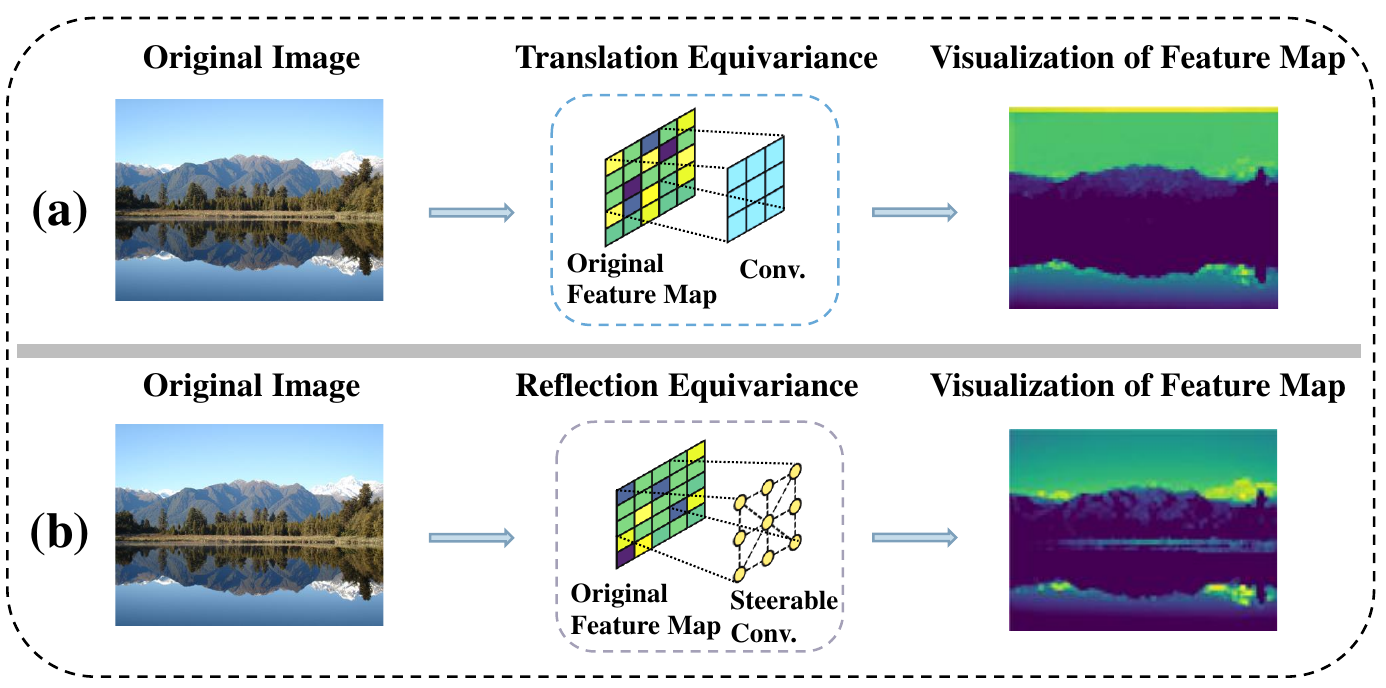}  
  \caption{(a) visualizes the feature map of a water reflection image extracted using traditional convolution with translational equivariance, while (b) visualizes the feature map extracted using steerable convolution with rotational and reflective equivariance. The visualization in (b) form a clearer representation.}
  \label{fig:ESCNN}
\end{figure}

\subsection{Preliminaries}
\subsubsection{Equivariant Learning}
Steerable convolution is a type of convolution whose feature space consists of steerable feature fields, which obey transformation laws associated with the group representations of $E(2)$ subgroups. As illustrated in Fig.~\ref{fig:ESCNN}, the upper panel displays the feature map of the water reflection image extracted by a conventional Convolutional Neural Network (CNN), while the lower panel presents the feature map extracted by a network constructed with steerable convolutions. It can be clearly observed from the figure that the network built with steerable convolutions yields feature maps with more distinct contours and more complete symmetric features.

The Euclidean group, also known as the rigid motion group or isometry group, constitutes the mathematical structure describing all distance-preserving transformations in the Euclidean space. These transformations include translations, rotations, reflections, and their arbitrary combinations. The Euclidean group is usually denoted as $E(n)$, where $n$ is the dimension of the Euclidean space. It consists of all transformations $f: \mathbb{R}^{n} \rightarrow \mathbb{R}^{n}$ that satisfy
\begin{equation}
\lVert f(x) - f(y) \rVert = \lVert x - y \rVert, \forall x, y \in \mathbb{R}^n,
\end{equation}
where $\left \|.\right \| $ is the Euclidean norm. Here we are mainly concerned with the group $E(n)$ of isometric transformations on the plane $\mathbb{R}^{2}$. The Euclidean group, $E(n)$, is composed of the translation group $\left(\mathbb{R}^{2},+\right)$ and the orthogonal group $O(2)=\left \{ O\in \mathbb{R}^{2\times2}|O^{T}O=id_{2\times2}\right \}$ via the semi-direct product operation, i.e.
\begin{equation}
E(2)\cong(\mathbb{R}^{2} ,+)\rtimes O(2). 
\end{equation}
We mainly consider subgroups of the form $(\mathbb{R}^{2} ,+) \rtimes G$, where $G\le O(2)$ \cite{weiler2019general}. 

The feature space of a traditional CNN is a collection of multi-channel feature maps that represent the image features at particular scales or frequencies. The weight-sharing property of CNN convolutional kernels ensures translation equivariance in the feature space, i.e., a translation of input features results in corresponding translated output features. 

In contrast, the feature space of a steerable CNN is the steerable feature field $ f: \mathbb{R}^{2} \rightarrow \mathbb{R}^{c} $, which associates a $c$-dimensional feature vector, $ f(x)\in \mathbb{R}^{c} $, with each point $x$ on the plane $ \mathbb{R}^{2} $, and is related to the transformation law of $E(n)$. For example, for vector eigenfields $ v:\mathbb{R}^{2}\to \mathbb{R}^{2} $ (as in a gradient image), the action of $E(n)$ is to move each pixel to a new position and also to change its direction by the action of $ g\in G $. While the transformation law of a general feature field $ f:\mathbb{R}^{2}\to \mathbb{R}^{c} $ is completely determined by its type $\rho$, the $\rho:G\mapsto GL(\mathbb{R}^{c}) $ of a steerable feature field is a group representation that specifies how the $c$ channels of each feature vector $f(x)$ are mixed in the feature space. Like the feature space of a conventional CNN that contains multiple channels, the feature space of a steerable CNN is composed of multiple feature fields $ f_{i}:\mathbb{R}^{2}\to \mathbb{R}^{c_{i}} $, each of which has its own type $\rho_{i}:G\to GL(\mathbb{R}^{c_{i}})$.

Steerable convolution \cite{weiler2019general} not only exhibits translation equivariance but also demonstrates equivariance to reflection and rotation. Therefore, the most general equivariant linear mapping between steerable feature spaces transforming under $\rho_{in}$ and $\rho_{out}$ is given by a $G$-steerable kernel $k:\mathbb{R}^{2}\to\mathbb{R}^{c_{out}\times c_{in} } $ that satisfies the kernel constraint \cite{weiler2019general}:
\begin{equation}
k(gx)=\rho_{out}(g)k(x)\rho_{in}(g^{-1}) \forall g\in G,x\in \mathbb{R}^{2}. 
\end{equation}
This constraint ensures the correct transformation behavior of input and output feature fields.

\subsubsection{Matrix Decomposition}
Matrix decomposition is a mathematical process that decomposes a given matrix into the product or sum of several matrices with specific mathematical properties. In the context of global context modeling, compared to self-attention mechanisms, matrix decomposition demonstrates relative advantages in computational efficiency and performance. From a mathematical perspective, matrix decomposition algorithms represent input images as linear combinations of basis images, achieving low-rank embedding reconstruction through submatrix decomposition of the input representation.

\begin{figure}[t]
  \centering
    \includegraphics[width=0.9\linewidth,keepaspectratio]{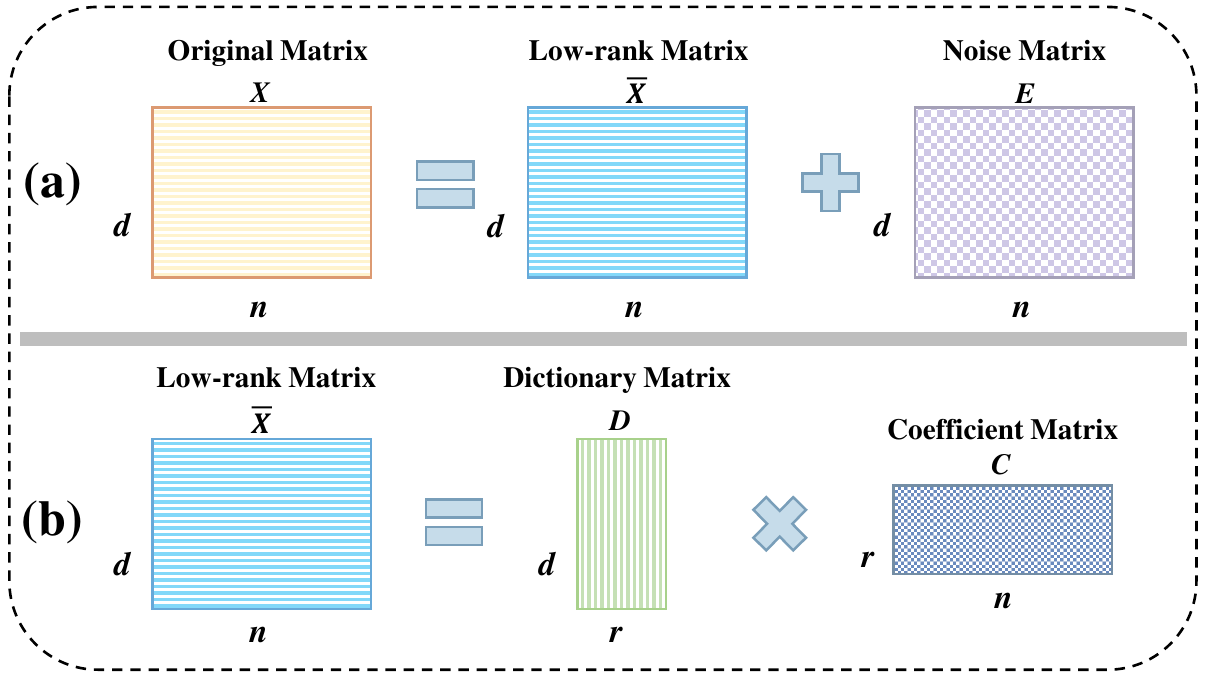} 
  \caption{Formulation of matrix decomposition. Specifically, (a) indicates that the original matrix is composed of the sum of a low-rank matrix and a noise matrix, while (b) illustrates that the low-rank matrix is constructed through the multiplication of a dictionary matrix and a coefficient matrix.}
  \label{fig:MD}
\end{figure}

If the image data are arranged into a large matrix 
\begin{equation}
X=\left [ x_{1},\dots,x_{n}\right]\in \mathbb{R}^{d\times n},
\end{equation}
where each column $ x_{i} $ corresponds to a pixel vector of an image, $d$ is the total number of pixels in the image, and $n$ is the number of images. It is often assumed that these image data imply a low-dimensional subspace structure, or consist of multiple subspaces. Specifically, there exist a dictionary matrix $ D $ and a coefficient matrix $ C $, i.e.,
\begin{equation}
\begin{split}
D &= \left [ d_{1},\dots,d_{r}\right]\in \mathbb{R}^{d \times r}, \\
C &= \left [ c_{1},\dots,c_{n}\right]\in \mathbb{R}^{r \times n}, \\
X &\approx DC,
\end{split}
\end{equation}
where the column vectors of the dictionary matrix $D$ can be treated as the base features of the images, and the coefficient matrix $C$ denotes the coefficients of the individual images that are linearly combined from these features \cite{geng2021attention}. As illustrated in Fig.~\ref{fig:MD}, this matrix decomposition process is depicted as:

\begin{equation}
\overline{X} = DC, \quad \text{where} \quad X = \overline{X} + E.
\end{equation}
$ \overline{X}\in \mathbb{R}^{d\times n} $ is the output low-rank reconstruction matrix, representing the main structural features of the image data, and $ E\in \mathbb{R}^{d\times n} $ is the noise matrix to be discarded, containing the high-frequency noises or non-primary features in the image. It is assumed here that the recovered matrix $ \overline{X} $ has the low-rank property, i.e., its rank satisfies 
\cite{geng2021attention}:
\begin{equation}
\small rank(\overline{X})\le min(rank(D),rank(C))\le r\ll min(d,n).
\end{equation}

In summary, the dictionary matrix, $D$, contains the basic features of the image; the coefficient matrix, $C$, reconstructs the image by combining these base features with additive noise, $E$. By making different assumptions about the structure, various matrix decomposition models can be derived for tasks such as feature extraction, dimensionality reduction, denoising, and image representation learning.

\begin{figure*}[t]
  \centering
  \includegraphics[width=1.0\linewidth,keepaspectratio]{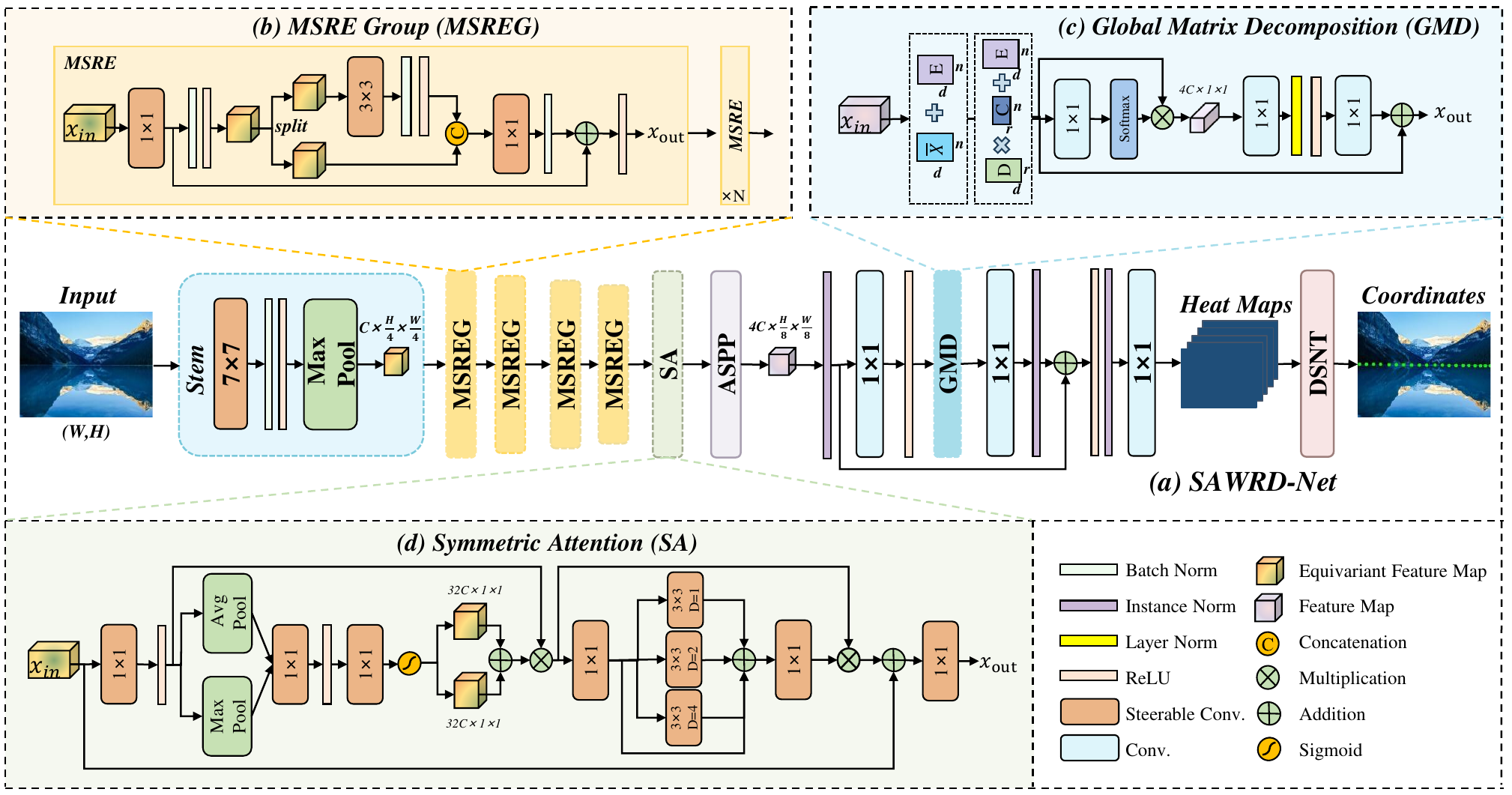}
  \caption{Illustration of the proposed water reflection detection network, SAWRD-Net. (a) depicts the overall architecture of the model. (b) shows the detailed structure of the proposed MSRE block. (c) illustrates the detailed structure of the Global Matrix Decomposition module within the MD decoder. (d) presents the detailed structure of the proposed symmetric attention mechanism.}
  \label{fig:SAWRD}
\end{figure*}


\subsection{Water Reflection Detection Network}

This section details the proposed Symmetry-Aware Water Reflection Detection Network. The overall architecture, illustrated in Fig.~\ref{fig:SAWRD}, is composed of three key components: the Multi-scale Reflection Equivariant block, the Symmetric Attention mechanism, and the Matrix Decomposition Decoder.

\subsubsection{Multi-scale Reflection Equivariant Convolutional Block}

Water reflection normally manifests an approximate mirror symmetry. However, the reflection patterns often exhibit discrete rotational or flipping deformations due to water surface fluctuations and variations in camera viewing angles and illumination directions during the imaging processes. To rigorously preserve the equivariance to these spatial transformations during feature extraction, while simultaneously balancing representation ability and computational efficiency, we employ $E(2)$-equivariant convolutions \cite{weiler2019general} based on the dihedral group, $D_{8}$‌, to construct a MSRE block, where the $D_{8}$ group‌ comprises eight discrete rotations (with angles as integer multiples of $\frac{2\pi}{8}$) and reflection transformations. We first perform a linear transformation along the channel dimension of input features through 1$\times$1 equivariant convolutions. 

Subsequently, the transformed feature tensor is partitioned into two mutually exclusive subspaces along the channel dimension. In the first subspace, we employ 3$\times$3 equivariant dilated convolution kernels that expand the receptive field while preserving group representation constraints, effectively capturing multi-scale contextual information. The second subspace retains original features to preserve low-level detail information. The processed outputs from both subspaces are then reintegrated through channel-wise concatenation, followed by a 1$\times$1 equivariant convolution to achieve unified fusion of multi-level features. 

This design paradigm simultaneously enables feature reuse for enhanced discriminative power and hierarchical extraction of multi-scale features through diversified receptive fields. Such architecture ensures effective multi-scale feature learning while strictly maintaining equivariance properties.

\subsubsection{Symmetric Attention Mechanism}

In general, the scale of water reflection varies while the reflective area often occupies only a small portion of the image. The intensity and morphology of the reflective area are influenced by external factors, such as illumination conditions, water quality, and surface fluctuations. Therefore, a detection network should be able to selectively enhance reflection-related characteristics while suppressing irrelevant background regions. To this end, we propose a composite Symmetric Attention (SA) mechanism inspired by the Equivariant Steerable Convolutional Neural Network (ESCNN) introduced by \citet{weiler2019general}‌. ESCNN uses channel attention \cite{woo2018cbam} and multi-scale spatial features to represent complex reflection patterns.

Our channel attention module utilizes a symmetric dual-path structure to extract global feature statistics. It applies Pointwise Adaptive Average Pooling and Pointwise Adaptive Max Pooling operations in parallel to capture both the average activation levels and the salient feature responses of the feature maps, thereby characterizing the channel-wise information distribution. 

The feature transformation is then performed by a two-layer shared parameter bottleneck structure. This structure first uses an equivariant convolution $1 \times 1$ to compress the channel dimension to $\frac{1}{4}$ of its original size, enhancing the interaction of features. After a non-linear activation with an Equivariant ReLU, another $1 \times 1$ equivariant convolution restores the original channel dimension. Finally, the channel attention weights are generated via a pointwise non-linear function, formally expressed as:
\begin{equation}\label{eq:eca}
\begin{split}
H_{n+1} = H_n \otimes \Big( \sigma \Big( 
    &\mathcal{W}_{2}\big(\delta (\mathcal{W}_{1}(\text{AvgPool}(H_n)))\big) \\
    &+ \mathcal{W}_{2}\big(\delta (\mathcal{W}_{1}(\text{MaxPool}(H_n)))\big) \Big) \Big),
\end{split}
\end{equation}
where $\otimes$ represents element-wise multiplication, $H$ represents the feature map, $H_n$ represents the feature map before input, and $H_{n+1}$ represents the feature map after input. $\mathcal{W}_{1}$ and $\mathcal{W}_{2}$ represent equivariant convolutions $1 \times 1$ for the reduction and expansion of the dimensionality, respectively, $\delta$ denotes the activation of ReLU and $\sigma$ represents the sigmoid function.

To capture multi-scale reflection features, a spatial feature extraction module with heterogeneous receptive fields is introduced. This module adopts a multi-branch architecture. The basic branch employs a standard $3 \times 3$ equivariant group convolution (dilation rate = 1) to capture fine-grained local details. Two multi-scale branches utilize $3 \times 3$ equivariant dilated convolutions with dilation rates of 2 and 4, respectively, to progressively expand the receptive field and model spatial context at different scales, and padding is used to ensure that the feature maps output by each branch have the same size. The outputs of all branches are aggregated via residual connections and fused using a $1 \times 1$ equivariant convolution. The resulting features are then multiplied element-wise with the original input features. 

\begin{figure*}[t]
 \centering{
  \includegraphics[width=1.0\linewidth,keepaspectratio]{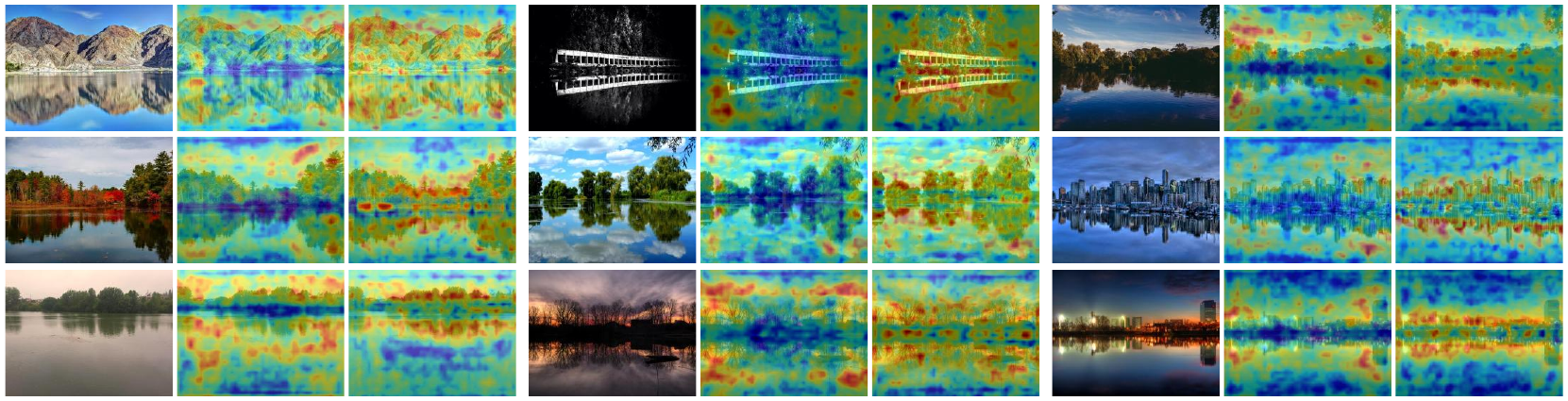}\par}
\vspace{-0.8em}
\begin{tabular}{@{}*{9}{>{\scriptsize}p{0.94cm}<{\centering\arraybackslash}}@{}}
Image  &
Before  &
After  & 
Image  &
Before  &
After & 
Image  &
Before  &
After
\end{tabular}
\vspace{-1.3 em}
  \caption{Visualization of attention maps. Here, column ``Image" represents the input images fed into the model, column ``Before" indicates the visualization of feature maps prior to the application of symmetric attention, and column ``After" shows the visualization of feature maps following the utilization of symmetric attention. The visualization results clearly demonstrate that our symmetric attention mechanism can effectively enhance the weights of the regions of interest.}
  \label{fig:Attention}
\end{figure*}

This multi-scale design enables the model to capture both local details and global distribution characteristics of water reflections, significantly improving its adaptability. As visualized in Fig.~\ref{fig:Attention}, this process effectively enhances the regions of interest. The operation can be formally expressed as
\begin{equation}\label{eq:msa}
\begin{split}
H_{n+1} = H_n \otimes \Biggl(
    \mathcal{W}_{c} \Biggl( \sum_{d \in \{1,2,4\}} \mathcal{W}_{d}(H_n) \Biggr) \Biggr),
\end{split}
\end{equation}
where $\otimes$ represents element-wise multiplication, $H$ represents the feature map, $H_n$ represents the feature map before input, and $H_{n+1}$ represents the feature map after input. $\mathcal{W}_{d}$ denotes the convolution operation with the dilation rate $d$, and $\mathcal{W}_{c}$ represents the feature fusion layer. This architecture explicitly establishes cross-scale representations, bridging pixel-level details with region-level context.


\subsubsection{‌Matrix Decomposition Decoder‌}
Global contextual dependency is decisive in water reflection detection, as most reflections exhibit long-range characteristics and are rendered as an imperfect reflective symmetry due to external factors such as water waves and illumination. Convolutional operations, while effective for local pattern extraction, inherently lack the ability to explicitly model these global spatial relationships. In contrast, matrix decomposition offers significant advantages for this task. Its intrinsic low-rank property effectively models the global contextual dependencies in water reflection images, and it is capable of eliminating the influence of noise through its decomposition process. 

Although the self-attention mechanism can also model global dependencies, it has two limitations in water reflection detection. First, the space complexity of it is $O(n^2 d)$ (where $n = H \times W$ is the number of spatial locations and $d$ is the number of channels). Second, it is susceptible to noise interference. In contrast, matrix decomposition has a space complexity of only $O(n d r)$ (where the low-rank rank $r \ll n$), and can explicitly separate noise from the global structure.


Therefore, our decoder is designed to model long-range dependencies and filter noise by building on the principles of matrix decomposition \cite{geng2021attention}. The process begins by preparing the features: we first apply Instance Normalization \cite{ulyanov2016instance} to standardize the statistics of each channel, followed by a $1 \times 1$ convolution with a ReLU activation to linearly map the features into a suitable embedding space. The core of the decoder is a Global Matrix Decomposition (GMD) block that performs low-rank signal recovery. This module takes the embedded feature maps and decomposes them into the product of a dictionary matrix and a coefficient matrix. 

Specifically, the features are modeled as the product of a dictionary and coefficients plus noise. A low-rank reconstruction is obtained by alternately updating the coefficients and the dictionary for six iterations using multiplicative update rules. The objective function converges linearly and stabilizes after six iterations. The computational complexity is $O(ndr)$. Each forward propagation takes around 15 milliseconds, and the memory usage is one-fifth that of the self-attention mechanism. This process is key to effectively extracting the essential global correlations from the scene while simultaneously suppressing the high-frequency noise caused by water surface distortions.



In the GMD block, after obtaining the low-rank signal subspace, this subspace is subsequently utilized to aggregate global contextual information. First, a $1 \times 1$ convolution and a softmax function are applied to the subspace representation to generate attention-like weights. These weights are then used in a pooling mechanism to produce a globally aggregated feature vector through weighted averaging. This vector represents a summary of the entire scene's global context. Next, this global context vector is refined and fused with the original local features. A bottleneck structure is employed to perform non-linear transformations on the aggregated vector, which captures complex inter-channel dependencies. These global contextual features are then fused with the original position-wise features through element-wise addition. This critical step enriches the local features with an understanding of the global scene, allowing the model to make more informed predictions.


Finally, an output module projects the fused features into the final heat maps. This begins with a linear transformation, using a $1 \times 1$ convolution and a ReLU activation, to reduce the parameter count. Subsequently, Instance Normalization \cite{ulyanov2016instance} is reapplied to stabilize the channel-wise statistics of the output, followed by a final $1 \times 1$ convolution for the channel-wise linear transformation that generates the heat maps. This overall decoder design excels at modeling the long-range dependencies inherent in water reflection. By explicitly extracting the underlying low-rank data structure, it demonstrates superior performance in water reflection detection tasks.


\section{Experimental Setup}
To rigorously validate the performance of our proposed SAWRD-Net and to dissect the contributions of its core architectural components, we designed a multi-faceted experimental protocol. Our evaluation is centered on the large-scale Water Reflection Scene Data set (WRSD) benchmark \cite{dong2024wrd}, where SAWRD-Net is compared against a comprehensive suite of state-of-the-art methods. We also conduct a series of detailed ablation studies to systematically isolate and quantify the impact of each architectural innovation. All other design choices and the training/evaluation protocol follow the setup defined earlier, enabling unambiguous attribution of each contribution.



\subsection{Data Set}
As the largest currently available data set for this task, WRSD \cite{dong2024wrd} contains 2,117 high-quality images, each with accurately manually annotated symmetry axis points. The data set is characterized by its diversity. Specifically, the symmetry axes are presented at various tilt angles, and the scenes encompass a rich variety of content with a wide range of scale variations, providing a comprehensive set of test samples.


\subsection{Evaluation Metric}
The primary evaluation metric is the True Positive (TP) Rate \cite{nagar2019reflection}, which considers both the angular deviation ($\Delta\theta$) and the midpoint distance ($\Delta d$) between the predicted and ground-truth axes. A prediction is classified as TP if its angular deviation is below a threshold $\theta$ and its distance deviation is below a threshold $d$; otherwise, it is marked as a False Positive (FP). The TP Rate is then calculated as:
\begin{equation}
\text{TP Rate} = \frac{\text{TP}}{\text{TP} + \text{FP}}.
\end{equation}
Following the settings in \cite{dong2024wrd}, we set the distance threshold $d$ to 2.5\% of the image height (i.e., $d=7.5$ for a 300-pixel tall image) and the angle threshold $\theta$ to $\frac{\pi}{60}$ radians (i.e., $\theta = 3^{\circ}$). To assess the stability and reliability of each method, we also report the mean and standard deviation of the angle and distance differences for all test images.


\subsection{Baselines}


To comprehensively and fairly evaluate SAWRD-Net, we compare it against five families of baseline methods: (i) a task-specific water reflection detector, WRD-Net~\cite{dong2024wrd}; (ii) four detectors for reflection or rotational-symmetry~\cite{funk2017beyond,seo2021learning,seo2022reflection}; (iii) ten strong dense-prediction or segmentation backbones~\cite{ronneberger2015u,bullock2018xnet,dai2021coatnet,wen2024crnet,yang2022focal,SunXLW19,yu2024inceptionnext,liu2024rolling,huang2022stvit,xie2021segformer}; (iv) five classical symmetry methods~\cite{loy2006detecting,cicconet2014mirror,cicconet2017finding,elawady2017wavelet,gnutti2021combining}; and (v) two typical object detectors~\cite{tian2025yolov12,sapkota2025yolo26}. We adapt each method to the water reflection \emph{axis-point} task using the authors’ released code where available, and compare predicted axes with the ground-truth data.

\subsection{Implementation Details}
The images are divided into two categories, in which the reflection axis is approximately horizontal or vertical. All images are fed into the network at the original resolution of $400 \times 300$ pixels. Data augmentation is not used during the training process. Model weights are saved at the end of each training epoch and early stopping is not employed. We use the Adam optimizer. The learning rate, weight decay and mini-batch size are set to $5{\times}10^{-4}$, $10^{-4}$ and $2$, respectively. The learning rate remains constant throughout the training process without a decay strategy. In total, each model is trained for 100 epochs. A three-fold \emph{image-level} cross-validation is conducted, which is identical to that used by WRD-Net~\cite{dong2024wrd}. Performance metrics are computed over all test images across the three folds using the same TP Rate definition and thresholds ($\theta=\pi/60$, $d=2.5\%$ of image height).

SAWRD-Net is implemented in Python~3.8.20 with PyTorch~1.11.0 and CUDA~11.5. All experiments are performed on a single NVIDIA RTX~3090 GPU. All baselines are implemented in PyTorch, with an identical DSNT\cite{nibali2018numerical} head appended to regress reflection-axis keypoints. Although PMCNet \cite{seo2021learning} and EquiSym \cite{seo2022reflection} provide pre-trained weights on symmetric data sets, the training on the WRSD \cite{dong2024wrd} can improve detection accuracy. To ensure fair comparison, we do not use pre-trained weights unless otherwise specified. That is, all baselines are trained \emph{from scratch} under a unified protocol.



\section{Experimental Results}
Following the settings detailed in Section 4, experiments were performed. This section elucidates the subsequent results.

\subsection{Comparison with State-of-the-Art Methods}

\begin{figure*}[t]
  \centering
  \includegraphics[width=\linewidth,keepaspectratio]{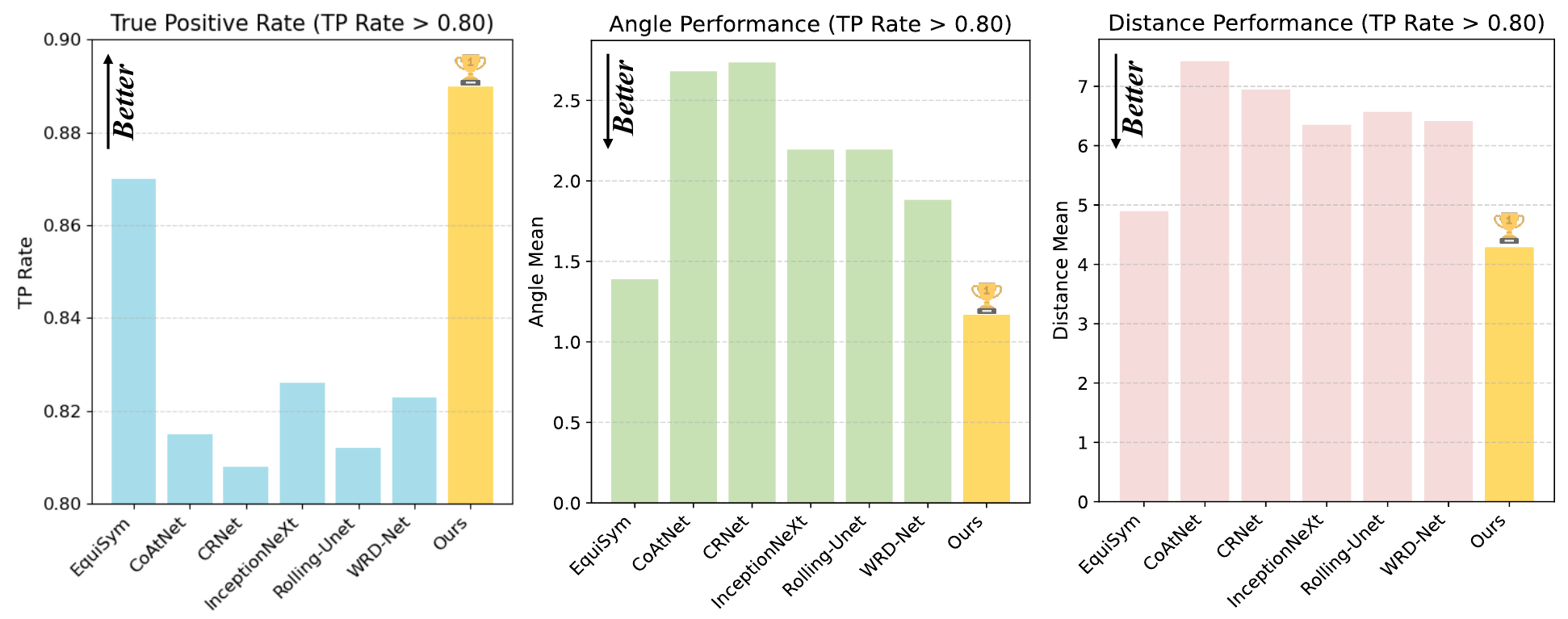}
  \caption{Visualization of comparison with state-of-the-art methods. We visualized models with a TP Rate greater than 0.80, demonstrating the superiority of our method across various evaluation metrics such as TP Rate, angle, and distance.}
  \label{fig:Chart1}
\end{figure*}

Table~\ref{tab:comparison} demonstrates the performance of the proposed SAWRD-Net against 22 state-of-the-art methods. Results marked with \textsuperscript{\dag} are quoted from \cite{dong2024wrd}; all other entries were obtained under the same experimental settings (Sec.~4-4). Under the standard thresholds ($\theta=\pi/60$, $d=2.5\%$ of image height), SAWRD-Net achieves a TP Rate~\cite{nagar2019reflection} of 0.890 and obtains the best angle and distance statistics in terms of mean and standard deviation. None of the baseline methods outperform SAWRD-Net. Fig.~\ref{fig:Chart1} visualizes the TP Rate, angle mean and distance mean values derived using the models which produce a TP Rate greater than 0.80. The results show that our method performs best across the three evaluation metrics.

\begin{table*}[t]
\renewcommand{\arraystretch}{1.0}
\centering
\setlength{\tabcolsep}{3.0pt}
\caption{Comparison between the baselines and SAWRD-Net in terms of different performance metrics.}

\label{tab:comparison}
{
\scriptsize
\begin{tabular}{c|c|ccc|ccc}
\hline
\multirow{2}{*}{\textbf{Method}}  & \multirow{2}{*}{\textbf{TP Rate \cite{nagar2019reflection}}} & \multicolumn{3}{c|}{\textbf{Angle}} & \multicolumn{3}{c}{\textbf{Distance}} \\
\cline{3-8}
 &  & \textbf{Mean} & \textbf{Var.} & \textbf{Std.} & \textbf{Mean} & \textbf{Var.} & \textbf{Std.} \\
\hline
Loy and Eklundh's\textsuperscript{$\dagger$} \cite{loy2006detecting} (ECCV 2006) & 0.682 & 4.478 & 274.684 & 16.574 & 41.270 & 48727.560 & 220.743 \\
Cicconet et al.'s\textsuperscript{$\dagger$} \cite{cicconet2014mirror} (CVPR 2014) & 0.512 & 32.000 & 1736.433 & 41.671 & 28.298 & 1435.399 & 37.887 \\
Cicconet et al.'s\textsuperscript{$\dagger$} \cite{cicconet2017finding} ICCV 2017) & 0.173 & 24.754 & 1194.133 & 34.556 & 38.372 & 1274.303 & 35.697 \\
Elawady et al.'s\textsuperscript{$\dagger$} \cite{elawady2017wavelet} (ICCV 2017) & 0.184 & 43.768 & 1936.014 & 44.000 & 26.120 & 697.579 & 26.412 \\
Gnutti et al.'s\textsuperscript{$\dagger$} \cite{gnutti2021combining} (TIP 2021) & 0.529 & 5.257 & 377.350 & 19.426 & 22.407 & 1169.404 & 34.197 \\
Sym-VGG \textsuperscript{$\dagger$} \cite{funk2017beyond} (ICCV 2017) & 0.330 & 11.138 & 807.850 & 28.423 & 19.866 & 437.702 & 20.921 \\
Sym-ResNet\textsuperscript{$\dagger$} \cite{funk2017beyond} (ICCV 2017) & 0.594 & 4.866 & 175.251 & 13.238 & 13.077 & 496.902 & 22.314 \\
PMCNet\textsuperscript{$\dagger$} \cite{seo2021learning} (ICCV 2021) & 0.670 & 5.533 & 228.725 & 15.124 & 12.664 & 589.244 & 24.274 \\
UNet\textsuperscript{$\dagger$} \cite{ronneberger2015u} (MICCAI 2015) & 0.461 & 5.860 & 224.266 & 14.976 & 18.830 & 576.884 & 24.018 \\
XNet\textsuperscript{$\dagger$} \cite{bullock2018xnet} (SPIE 2019) & 0.234 & 9.160 & 301.653 & 17.368 & 30.607 & 727.304 & 26.969 \\
EquiSym \cite{seo2022reflection} (CVPR 2022) & 0.870 & 1.389 & 61.655 & 7.852 & 4.898 & 175.530 & 13.248 \\
CoAtNet \cite{dai2021coatnet} (NeurIPS 2021) & 0.815 & 2.681 & 158.055 & 12.572 & 7.423 & 288.507 & 16.985 \\
CRNet \cite{wen2024crnet} (JKSUCI 2024) & 0.808 & 2.736 & 133.182 & 11.540 & 6.948 & 292.675 & 17.107 \\
FocalNet \cite{yang2022focal} (NeurIPS 2022) & 0.772 & 3.317 & 166.455 & 12.901 & 7.852 & 273.159 & 16.527 \\
HRNet \cite{SunXLW19} (CVPR 2019) & 0.795 & 2.706 & 120.416 & 10.973 & 7.763 & 328.204 & 18.116 \\
InceptionNeXt \cite{yu2024inceptionnext} (CVPR 2024) & 0.826 & 2.194 & 108.900 & 10.435 & 6.353 & 215.829 & 14.691 \\
Rolling-Unet \cite{liu2024rolling} (AAAI 2024) & 0.812 & 2.195 & 101.183 & 10.058 & 6.568 & 238.807 & 15.453 \\
STViT \cite{huang2022stvit} (CVPR 2023) & 0.378 & 8.667 & 570.263 & 23.880 & 18.088 & 442.536 & 21.036 \\

 YOLOv12 \cite{tian2025yolov12} (NeurIPS 2025)  &  0.785  & 2.802 & 125.893 & 11.220 & 7.763 & 266.749 & 16.332 \\

 YOLO26 \cite{sapkota2025yolo26} (arXiv 2025)  &  0.794  & 3.108 & 154.023 & 12.410 & 7.358 & 255.281 & 15.977 \\

 SegFormer \cite{xie2021segformer} (NeurIPS 2021)  &  0.798  & 2.008 & 124.696 & 11.700 & 6.373 & 251.449 & 15.857 \\

WRD-Net \textsuperscript{$\dagger$}\cite{dong2024wrd} (PR 2024) & 0.823 & 1.882 & 81.395 & 9.022 & 6.413 & 250.276 & 15.820 \\ 
\hline
SAWRD-Net (Ours) & \textbf{0.890} & \textbf{1.168} & \textbf{52.916} & \textbf{7.272} & \textbf{4.287} &\textbf{147.818} & \textbf{12.158} \\
\hline
\end{tabular}
}
\raggedright\scriptsize The results of the baselines marked with \textsuperscript{$\dagger$} were directly obtained from \cite{dong2024wrd}.
\end{table*}

Deep learning methods generally surpass traditional symmetry detectors on WRSD. Traditional approaches often assume near-perfect symmetry and degrade under the imperfect symmetry of natural water reflections, which are affected by ripples, refraction, and scattering. For dense-prediction baselines, we evaluate strong image-segmentation models because our task involves regressing 20 points along the reflection axis and thus relates to pixel-level prediction. Although these backbones perform well at pixel classification, they lack explicit symmetry priors and therefore yield lower TP Rates~\cite{nagar2019reflection}. 

In terms of the two state-of-the-art YOLO-based detectors, i.e., YOLOv12 \cite{tian2025yolov12} and YOLO26 \cite{sapkota2025yolo26}, we discarded the detection head that generates prediction bounding boxes but using the DSNT \cite{nibali2018numerical} head to predict points of the symmetric axis. Since YOLO-based methods were originally designed for sparse bounding box detection, they cannot produce high-resolution dense features or perform explicit global geometric modeling.

Among reflection-symmetry detection models, EquiSym~\cite{seo2022reflection} performs best with a TP Rate of 0.870, which is inferior to SAWRD-Net. This gap reflects the difference between idealized mirror symmetry targeted by generic symmetry detectors and the imperfect symmetry present in water scenes. By encoding symmetry in the encoder and suppressing noise via matrix decomposition in the decoder to recover low-rank structure, SAWRD-Net achieves the top TP Rate value. Compared to WRD-Net \cite{dong2024wrd}, SAWRD-Net introduces a symmetric attention mechanism and equivariant convolutions to explicitly model reflection symmetry, while replacing computationally intensive ViT branches with the efficient MSRE groups and matrix decomposition decoder, thus reducing computational cost.

\subsection{Qualitative Results}

\begin{figure*}[t]
  \centering{
  \includegraphics[width=1.0\linewidth,keepaspectratio]{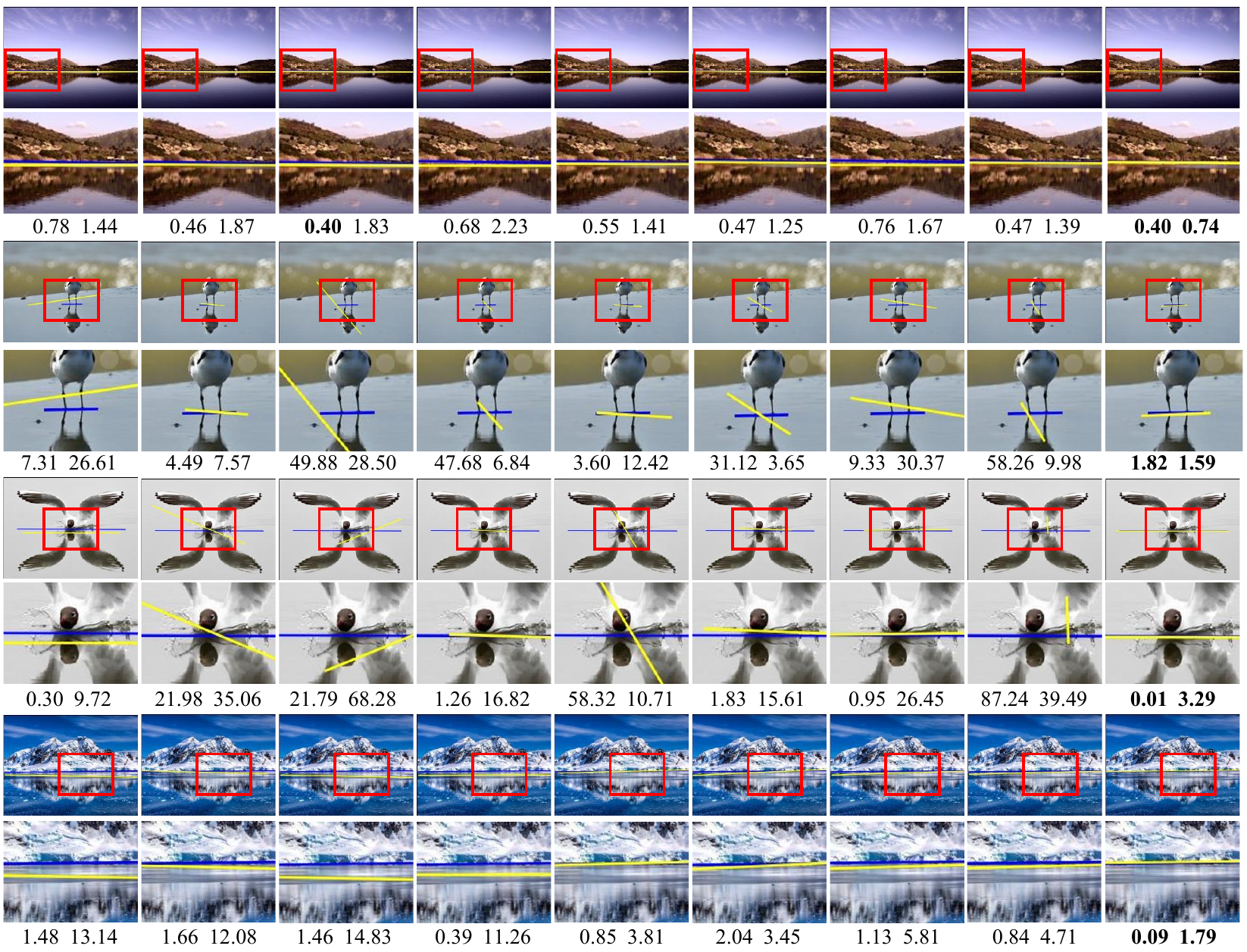}\par}
\vspace{-0.8em}
\begin{tabular}{@{}*{9}{>{\tiny}p{0.95cm}<{\centering\arraybackslash}}@{}}
HRNet \cite{SunXLW19} &
FocalNet \cite{yang2022focal} &
CRNet \cite{wen2024crnet} &
CoAtNet \cite{dai2021coatnet} &  
EquiSym \cite{seo2022reflection} & 
Rolling-Unet \cite{liu2024rolling} &
InceptionNeXt \cite{yu2024inceptionnext} &  
WRD-Net \cite{dong2024wrd} &
Ours
\end{tabular}
\vspace{-0.8em}
\caption{Qualitative results of water reflection detection on the WRSD. Within each group of the figure, the upper row  shows the detection results of eight baselines and SAWRD-Net, while the lower row shows the zoomed-in view of a selected region in the corresponding image. The yellow lines represent the axes detected by the models, while the blue lines denote the ground-truth. The two values displayed below an image indicate the angle and distance calculated between the detected axis and the ground-truth axis.}

\label{fig:Visual}
\end{figure*}

\begin{figure*}[t]
  \centering{
  \includegraphics[width=1.0\linewidth,keepaspectratio]{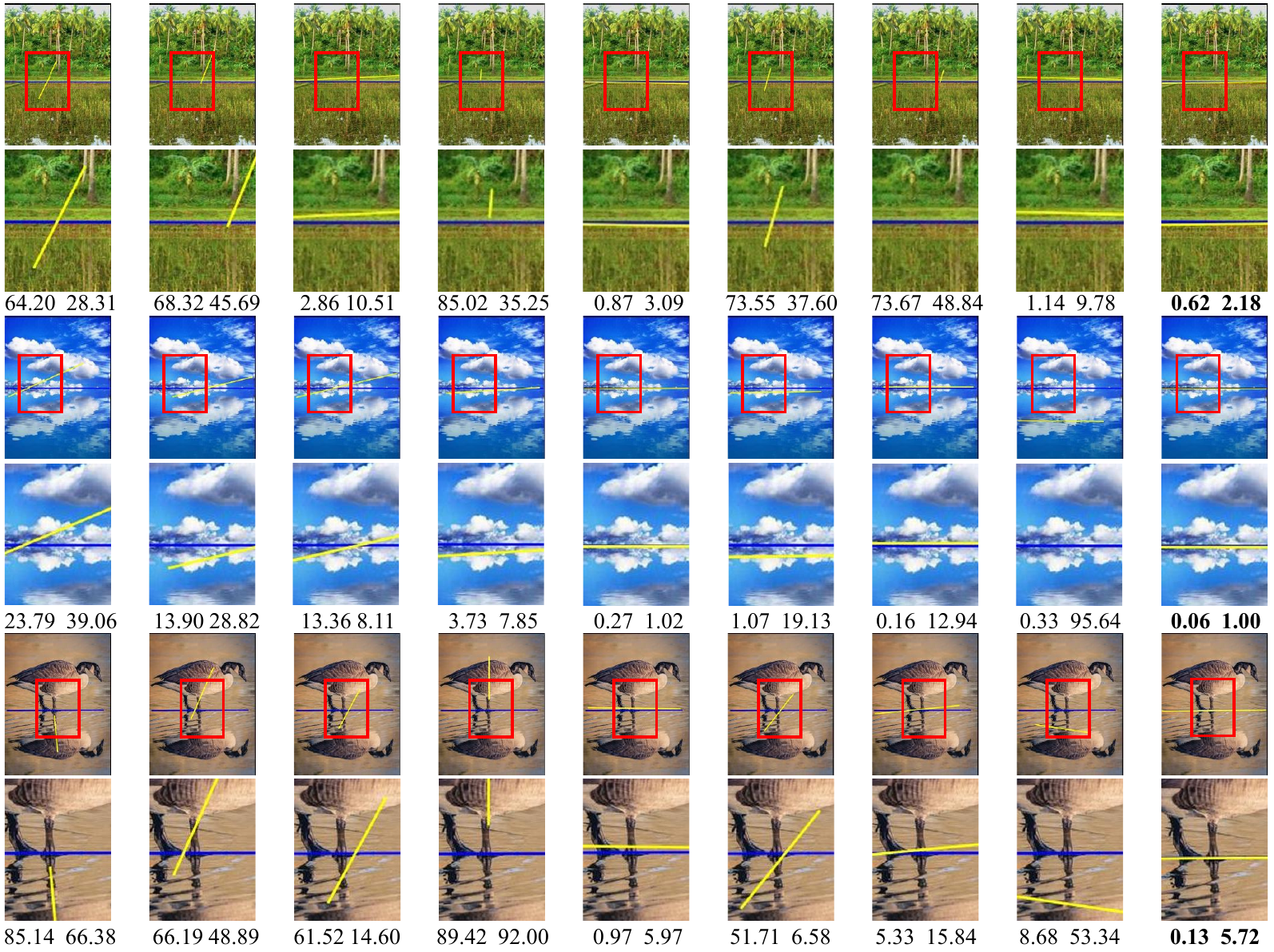}\par}
\vspace{-0.8em}
\begin{tabular}{@{}*{9}{>{\tiny}p{0.95cm}<{\centering\arraybackslash}}@{}}
HRNet \cite{SunXLW19} &
FocalNet \cite{yang2022focal} &
CRNet \cite{wen2024crnet} &
CoAtNet \cite{dai2021coatnet} &  
EquiSym \cite{seo2022reflection} & 
Rolling-Unet \cite{liu2024rolling} &
InceptionNeXt \cite{yu2024inceptionnext} &  
WRD-Net \cite{dong2024wrd} &
Ours
\end{tabular}
\vspace{-0.8em}
  \caption{Qualitative results of water reflection detection on the WRSD. Note that the images displayed here are those used in the experiment after being rotated clockwise by 90°. Within each group of the figure, the upper row  shows the detection results of eight baselines and SAWRD-Net, while the lower row shows the zoomed-in view of a selected region in the corresponding image. The yellow lines represent the axes detected by the models, while the blue lines denote the ground-truth. The two values displayed below an image indicate the angle and distance calculated between the detected axis and the ground-truth axis. 
  }
  \label{fig:Visual1}
\end{figure*}



The qualitative comparison between SAWRD-Net and eight baseline methods (HRNet~\cite{SunXLW19}, FocalNet~\cite{yang2022focal}, CRNet~\cite{wen2024crnet}, CoAtNet~\cite{dai2021coatnet}, EquiSym~\cite{seo2022reflection}, Rolling-Unet~\cite{liu2024rolling}, InceptionNeXt~\cite{yu2024inceptionnext}, and WRD-Net~\cite{dong2024wrd}) is shown in Figs.~\ref{fig:Visual} and ~\ref{fig:Visual1}. These baselines were selected based on their top TP Rate rankings in Table~\ref{tab:comparison}. Figure~\ref{fig:Visual} displays the original experimental images, whereas Fig.~\ref{fig:Visual1} presents the same images rotated $90^\circ$ clockwise to aid visualization.

Across the examples, SAWRD-Net exhibits closer alignment with the ground-truth annotations and consistently smaller angular and spatial deviations than the compared methods. In challenging scenarios involving water waves or blurred reflective surfaces, SAWRD-Net maintains robust performance relative to the baselines. Taken together, these observations indicate that SAWRD-Net achieves higher detection accuracy and robustness than the existing baseline methods.

\subsection{Robustness to Different Interferences}

\begin{figure*}[t]
  \centering  \includegraphics[width=1.0\linewidth,keepaspectratio]{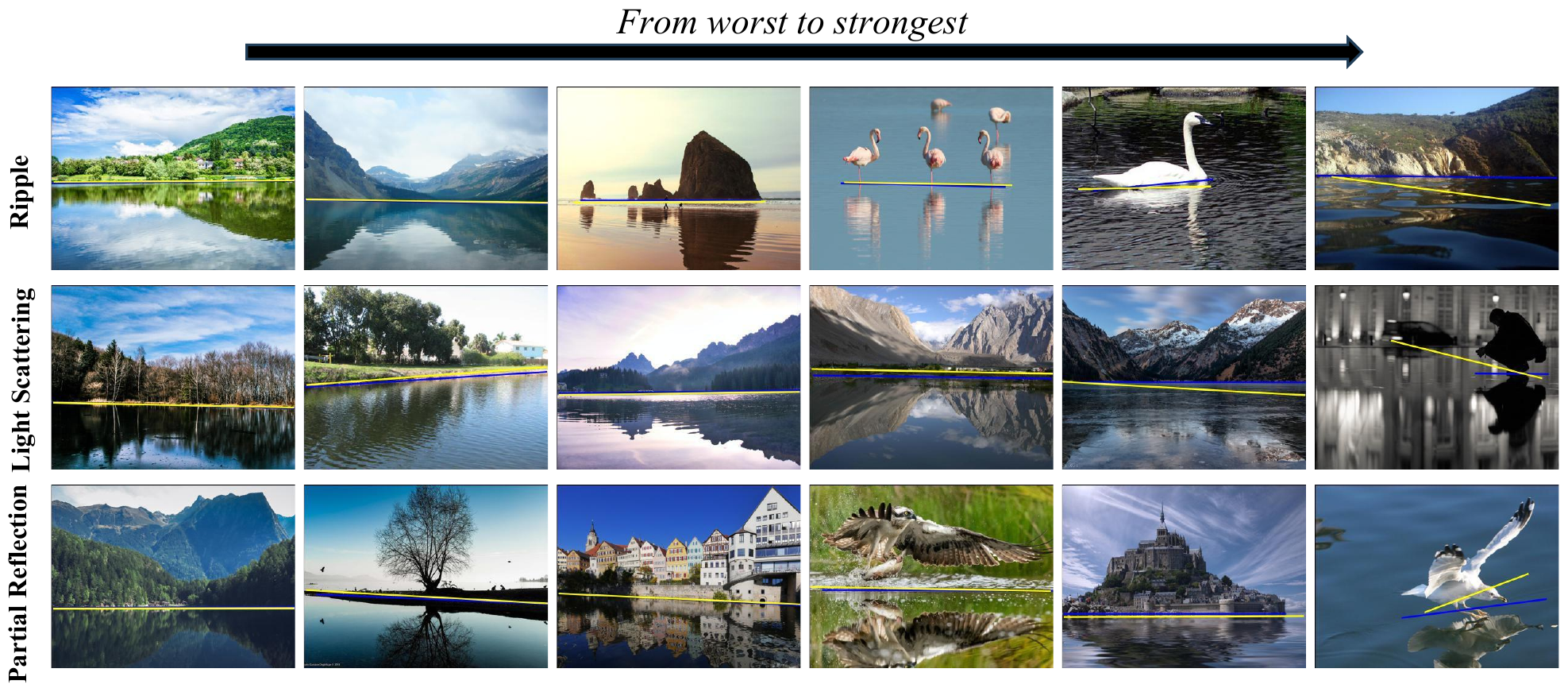}
  \caption{Examples of the water reflectance images suffered from three typical interferences, including ripples, light scattering, and partial reflections. Within each group, the degree of the interference increases from the left to the right.}
  \label{fig:CategoryAnalysis}
\end{figure*}

Despite achieving the highest TP Rate value, SAWRD-Net may encounter performance degradation when different interferences occur, such as ripples, light scattering, and partial reflections. To evaluate the robustness of SAWRD-Net to these interferences, we first randomly selected 100 images per interference type from the WRSD \cite{dong2024wrd}. In total, three subsets were obtained. Given a subset, six example images are shown in Fig.~\ref{fig:CategoryAnalysis}. SAWRD-Net was then applied to each subset. In terms of the ripple, light scattering, and partial reflection interferences, the TP Rate values obtained were 0.85, 0.91 and 0.87, respectively. As displayed in Fig.~\ref{fig:CategoryAnalysis}, SAWRD-Net can still detect the axis of reflectance in many challenging scenarios interfered by ripples, light scattering, and partial reflections. These results demonstrate that SAWRD-Net is robust to these interferences.

\subsection{Failure Cases}

\begin{figure*}[t]
  \centering
  \includegraphics[width=1.0\linewidth,keepaspectratio]{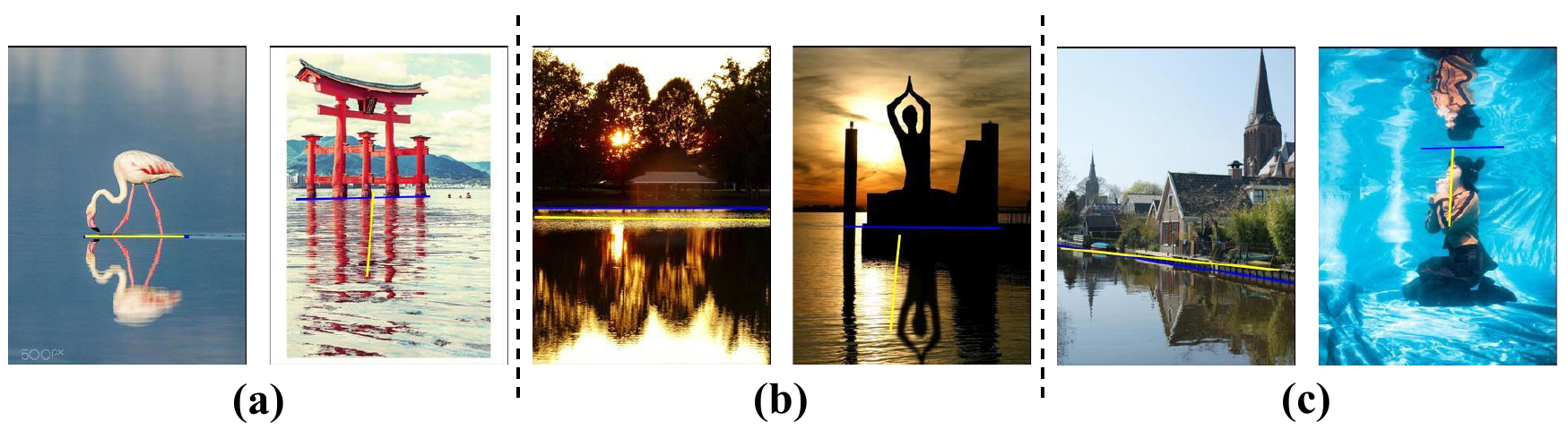}
  \caption{With regard to each of the three interferences, two failure cases are shown, in which (a) shows the detection results for different degrees of ripples; (b) presents the detection results under different degrees of light scattering; and (c)  displays the detection results for different degrees of partial reflections.}
  \label{fig:fail}
\end{figure*}

In terms of the three interferences, three sets of failure cases are shown in Fig.~\ref{fig:fail}, respectively. Within each set, two images are shown with progressively increasing interference degrees. For ripple interference, the Canny edge detector \cite{canny1986computational} and the Hough transform \cite{duda1972use} are used to extract ripple edges, and the degree of ripples is characterized by the range of the midpoint positions of the edges. Weak ripples (142 pixels) cause only minor deviations, while strong ripples (232 pixels) lead to larger detection errors. Regarding light scattering interference, the impact of illumination is evaluated using the ratio of saturated pixels and Sobel-based texture energy. Both overexposure and underexposure affect detection accuracy. However, weak illumination (only 5\% saturated pixels) reduces texture and contrast, resulting in poor detection performance. For partial reflection interference, the threshold-based binarization and connected component analysis are adopted to analyze the degree of partial reflections. Low partial reflections (largest connected component ratio of 92\%, 5 fragments) still yield coarse detection results, while high partial reflections (ratio of 30.5\%, 18 fragments) may lead to a complete failure.

To address the above-mentioned failures, future improvements could be implemented as follows. Deformable convolutions can be introduced for strong ripple deformations. Graph attention mechanisms can be applied to overcome fragmented reflections. A self-calibrated illumination module or frequency-domain features can be used to alleviate extreme lighting conditions. These solutions have potential to improve the detection stability of our SAWRD-Net within an end-to-end framework.

\subsection{Performance Analysis}

As shown in Table~\ref{tab:compare_flops_params_time}, SAWRD-Net achieves a good balance among computational complexity, number of parameters, and inference speed. Compared to WRD-Net~\cite{dong2024wrd}, it reduces FLOPs by around 31\% and also improves inference speed, with a moderate number of parameters. In particular, the GPU memory usage is around 587 MB during the inference stage while the peak memory usage (with $batchsize=2$) is around 2.9 GB in the training process. In general, SAWRD-Net demonstrates low computational complexity and moderate memory requirements, making it suitable for practical deployment.

\begin{table}[t]
\renewcommand{\arraystretch}{1.0}
\centering
\setlength{\tabcolsep}{9.0pt}
\caption{Comparison between the baselines and our SAWRD-Net in terms of computational complexity (FLOPs), number of parameters and average inference time (S).}
\label{tab:compare_flops_params_time}
{
\scriptsize
\begin{tabular}{c|cc|c}
\hline
\textbf{Method} & \textbf{FLOPs (G)} & \textbf{Params (M)} & \textbf{Average Inference Time (S)} \\
\hline
Loy and Eklundh's\textsuperscript{$\dagger$} \cite{loy2006detecting} (ECCV 2006) & N/A   & N/A   & 0.2551     \\
Cicconet et al.'s\textsuperscript{$\dagger$} \cite{cicconet2014mirror} (CVPR 2014) & N/A   & N/A   & 121.5111   \\
Cicconet et al.'s\textsuperscript{$\dagger$} \cite{cicconet2017finding} (ICCV 2017) & N/A   & N/A   & 0.4110     \\
Elawady et al.'s\textsuperscript{$\dagger$} \cite{elawady2017wavelet} (ICCV 2017) & N/A   & N/A   & 4.7615     \\
Gnutti et al.'s\textsuperscript{$\dagger$} \cite{gnutti2021combining} (TIP 2021) & N/A   & N/A   & 1079.8299  \\
Sym-VGG \textsuperscript{$\dagger$} \cite{funk2017beyond} (ICCV 2017) & 90.98 & 37.87 & 0.0126     \\
Sym-ResNet\textsuperscript{$\dagger$} \cite{funk2017beyond} (ICCV 2017) & 156.20& 19.95 & 0.0456     \\
PMCNet\textsuperscript{$\dagger$} \cite{seo2021learning} (ICCV 2021) & 227.62& 16.76 & 0.0322     \\
UNet\textsuperscript{$\dagger$} \cite{ronneberger2015u} (MICCAI 2015) & 56.83 & 13.40 & 0.0107     \\
XNet\textsuperscript{$\dagger$} \cite{bullock2018xnet} (SPIE 2019) & 95.41 & 26.64 & 0.0155     \\

EquiSym \cite{seo2022reflection} (CVPR 2022) & 42.15 & 16.44 & 0.0239     \\
CoAtNet \cite{dai2021coatnet} (NeurIPS 2021) & 12.85 & 17.87 & 0.0095     \\
CRNet \cite{wen2024crnet} (JKSUCI 2024) & 18.99 & 36.51 & 0.0267     \\
FocalNet \cite{yang2022focal} (NeurIPS 2022) & 109.45 & 59.84 & 0.0213     \\
HRNet \cite{SunXLW19} (CVPR 2019) & 43.26 & 65.40 & 0.0372     \\
InceptionNeXt \cite{yu2024inceptionnext} (CVPR 2024) & 46.58 & 86.69 & 0.0139     \\
Rolling-Unet \cite{liu2024rolling} (AAAI 2024) & 4.80 & 1.78 & 0.0214     \\
STViT \cite{huang2022stvit} (CVPR 2023) & 14.47 & 25.52 & 0.0213     \\
YOLOv12 \cite{tian2025yolov12} (NeurIPS 2025) & 0.92 & 2.14 & 0.0112     \\
YOLO26 \cite{sapkota2025yolo26} (arXiv 2025)  & 0.88 & 2.27 & 0.0089     \\
SegFormer \cite{xie2021segformer} (NeurIPS 2021) & 3.45 & 3.72 & 0.0078     \\
WRD-Net \textsuperscript{$\dagger$}\cite{dong2024wrd} (PR 2024) & 74.06 & 19.03 & 0.0789     \\
\hline
SAWRD-Net (Ours) & 51.11 & 23.97 & 0.0434 \\
\hline
\end{tabular}
}
\raggedright\scriptsize The results of the baselines marked with \textsuperscript{$\dagger$} were directly obtained from \cite{dong2024wrd}.
\end{table}

\subsection{Ablation Studies}
Unless otherwise noted, each ablation experiment varies a single factor under the unified protocol in Section 4.4. We report the TP Rate as defined in \cite{nagar2019reflection}.

\subsubsection{\(E(2)\)-Equivariant Convolutions}
The evaluation demonstrates the efficacy of equivariant learning for water reflection detection. We conducted a controlled comparison between models implemented with ESCNN~\cite{weiler2019general} and conventional CNN architectures, keeping all configurations identical except for the convolution operation. As shown in Table~\ref{tab:ec}, the conventional CNN model achieved a TP Rate~\cite{nagar2019reflection} of 0.797, whereas the equivariant-convolution variant attained 0.890. These results substantiate that adopting \(E(2)\)-equivariant convolutions positively impacts performance, with the complete architecture yielding optimal detection capability.

\begin{table}
\renewcommand{\arraystretch}{1.0}
\begin{center}
\normalsize
\setlength{\tabcolsep}{12pt}
\caption{TP Rate with and without equivariant convolution for training.}
\label{tab:ec}
\begin{tabular}{c|c}
\hline
Design & TP Rate \cite{nagar2019reflection} \\
\hline
w/o Equiv. & 0.797 \\
\hline
w/ Equiv. & \textbf{0.890} \\
\hline
\end{tabular}
\end{center}
\end{table}


\subsubsection{Dihedral Group}

To evaluate the effectiveness of the ESCNN~\cite{weiler2019general} with different dihedral groups used by our SAWRD-Net, we conducted a comparative experiment, which involves two models implemented based on the $D_4$ and $D_8$ groups, respectively. The other configurations were kept unchanged. As shown in Table~\ref{tab:dg}, the $D_4$-based model achieved a TP Rate value of 0.888, which is slightly lower than that derived using the model with $D_8$. This finding demonstrates that the choice of $D_8$ is reasonable.

\begin{table}
\renewcommand{\arraystretch}{1.0}
\begin{center}
\normalsize
\setlength{\tabcolsep}{12pt}
\caption{TP Rate with different dihedral groups during the training stage.}
\label{tab:dg}
\begin{tabular}{c|c}
\hline
Design & TP Rate \cite{nagar2019reflection} \\
\hline
w/ $D_{4}$ & 0.888 \\
\hline
w/ $D_{8}$ & \textbf{0.890} \\
\hline
\end{tabular}
\end{center}
\end{table}

\subsubsection{Effect of Symmetric Attention}
We conduct an ablation of the SA mechanism inserted between the backbone and the ASPP module~\cite{chen2017deeplab} under identical experimental settings. As shown in Table~\ref{tab:sa}, the model without SA achieves a TP Rate~\cite{nagar2019reflection} of 0.878, whereas enabling SA improves the TP Rate to 0.890, yielding an absolute gain of +0.012. These results indicate that SA contributes positively to performance, supporting its inclusion in the complete architecture.

\begin{table}
\renewcommand{\arraystretch}{1.0}
\begin{center}
\normalsize
\setlength{\tabcolsep}{12pt}
\caption{TP Rate with and without symmetric attention for training.}
\label{tab:sa}
\begin{tabular}{c|c}
\hline
Design & TP Rate \cite{nagar2019reflection} \\
\hline
w/o SA & 0.878 \\
\hline
w/ SA & \textbf{0.890} \\
\hline
\end{tabular}
\end{center}
\end{table}

\subsubsection{Matrix-Decomposition Decoder}
To evaluate the performance of the proposed Matrix Decomposition Decoder, we conducted targeted ablation experiments. We first compared the MD decoder with the Hamburger decoder~\cite{geng2021attention}. Although Hamburger has demonstrated strong performance as a general-purpose decoder in segmentation tasks, we observed limitations on the specific task of water reflection detection. Building on the general architectural pattern of Hamburger, we introduced an MD decoder tailored to the characteristics of water reflection detection. As shown in Table~\ref{tab:decoder}, using the standard Hamburger yielded a TP Rate~\cite{nagar2019reflection} of 0.878, whereas employing the MD decoder improves the TP Rate to 0.890 (absolute gain +0.012), validating the effectiveness of the proposed design.

Subsequently, to further examine the decoding advantages of the matrix-decomposition mechanism, we compared the MD decoder with the EquiSym decoder under a fixed encoder. Notably, EquiSym is a fully symmetric architecture constructed entirely from equivariant convolutions; since our encoder also employs equivariant convolutions, this comparison isolates the impact of the decoder architecture while keeping the encoder unchanged. As reported in Table~\ref{tab:md}, the EquiSym decoder attains a TP Rate of 0.879, whereas the MD decoder achieves 0.890 (absolute gain +0.011). These results confirm the advantage of matrix decomposition for handling water reflection features and also suggest synergistic effects between the network’s symmetry-aware encoder and the proposed decoder.

\begin{table}
\renewcommand{\arraystretch}{1.0}
\begin{center}
\normalsize
\setlength{\tabcolsep}{12pt}
\caption{TP Rate with hamburger or our decoder for training.}
\label{tab:decoder}
\begin{tabular}{c|c}
\hline
Module & TP Rate \cite{nagar2019reflection} \\
\hline
w/ Hamburger & 0.878 \\
\hline
w/ Our Decoder & \textbf{0.890} \\
\hline
\end{tabular}
\end{center}
\end{table}

\begin{table}
\renewcommand{\arraystretch}{1.0}
\begin{center}
\normalsize
\setlength{\tabcolsep}{12pt}
\caption{Impact of matrix decomposition on water reflection detection.}
\label{tab:md}
\begin{tabular}{c|c}
\hline
Design & TP Rate \cite{nagar2019reflection} \\
\hline
w/o MD & 0.879 \\
\hline
w/ MD & \textbf{0.890} \\
\hline
\end{tabular}
\end{center}
\end{table}

\section{Conclusion}
We presented SAWRD-Net, a symmetry-aware network for water reflection detection that integrates $E(2)$-equivariant learning with matrix decomposition. Building on symmetry-detection principles, SAWRD-Net leverages the imperfect reflective symmetry of natural water scenes by using equivariant convolutions and a symmetric attention mechanism to extract geometry-consistent, discriminative features and emphasize reflection-relevant regions. Treating axis localization as a structured modeling problem, we adopted a matrix decomposition decoder to capture global context and suppress noise, yielding a complementary synergy with the encoder. On the WRSD benchmark, SAWRD-Net attained state-of-the-art performance (TP Rate $0.890$ at $\theta=\pi/60$, $d=2.5\%$ of image height), and ablation experiments verified the contributions of equivariant features, SA placement, and the MD design. However, SAWRD-Net struggles with some challenging scenarios, such as severe ripples, extreme illumination, and fragmented reflections. In our future work, we will manage to address these challenges.

\section*{CRediT authorship contribution statement}
\textbf{Shuxuan Yao:} Data curation, Formal analysis, Methodology, Software, Validation, Visualization, Writing - original draft;
\textbf{Chengjia Wang:} Formal analysis, Methodology, Writing - Review \& Editing;
\textbf{Jianyuan Sun:} Writing -- review \& editing;
\textbf{Junyu Dong:} Resources, Funding acquisition;
\textbf{Xinghui Dong:} Conceptualization, Funding acquisition, Investigation, Methodology, Project Administration, Resources, Supervision, Writing - Review \& Editing.

\section*{Declaration of competing interest}
The authors declare that they have no known competing financial interests or personal relationships that could have appeared to influence the work reported in this paper.

\section*{Acknowledgement}
This study was supported in part by the National Natural Science Foundation of China (NSFC) (No. 42576200) and in part by the Key Research and Development Program of Shandong Province, China (No. 2024ZLGX06)

\section*{Data availability}
Data will be made available on request.

\bibliographystyle{elsarticle-num-names}
\bibliography{sawrd-net}



\end{document}


\begin{frontmatter}

\title{Supplementary Material for ``Water Reflection Detection Using Symmetric Attention''}

\author[label1]{Shuxuan Yao}
\ead{yaoshuxuan@stu.ouc.edu.cn}

\author[label2]{Chengjia Wang}
\ead{chengjia.wang@hw.ac.uk}

\author[label3]{Jianyuan Sun}
\ead{jianyuan.sun@dmu.ac.uk}

\author[label1]{Junyu Dong}
\ead{dongjunyu@ouc.edu.cn}

\author[label1]{Xinghui Dong\corref{cor1}}
\ead{xinghui.dong@ouc.edu.cn}

\cortext[cor1]{Corresponding author}

\affiliation[label1]{
  organization={State Key Laboratory of Physical Oceanography and the Faculty of Information Science and Engineering, Ocean University of China},
  addressline={238 Songling Road},
  city={Qingdao},
  postcode={266100},
  state={Shandong},
  country={China}
}

\affiliation[label2]{organization={School of Mathematical and Computer Sciences, Heriot-Watt University},
             addressline={Riccarton Mains Road, EH14 4AS, Edinburgh},
             country={United Kingdom}}

\affiliation[label3]{organization={School of Computer Science and Informatics, De Montfort University},
             addressline={Leicester, LE1 9BH},
             country={United Kingdom}}

\begin{abstract}
This supplementary material provides additional experimental results obtained using the proposed SAWRD-Net.
\end{abstract}

\end{frontmatter}

\section{Cross-Data-Set Validation}

\begin{figure}[ht]
  \centering
  \includegraphics[width=1.0\linewidth,keepaspectratio]{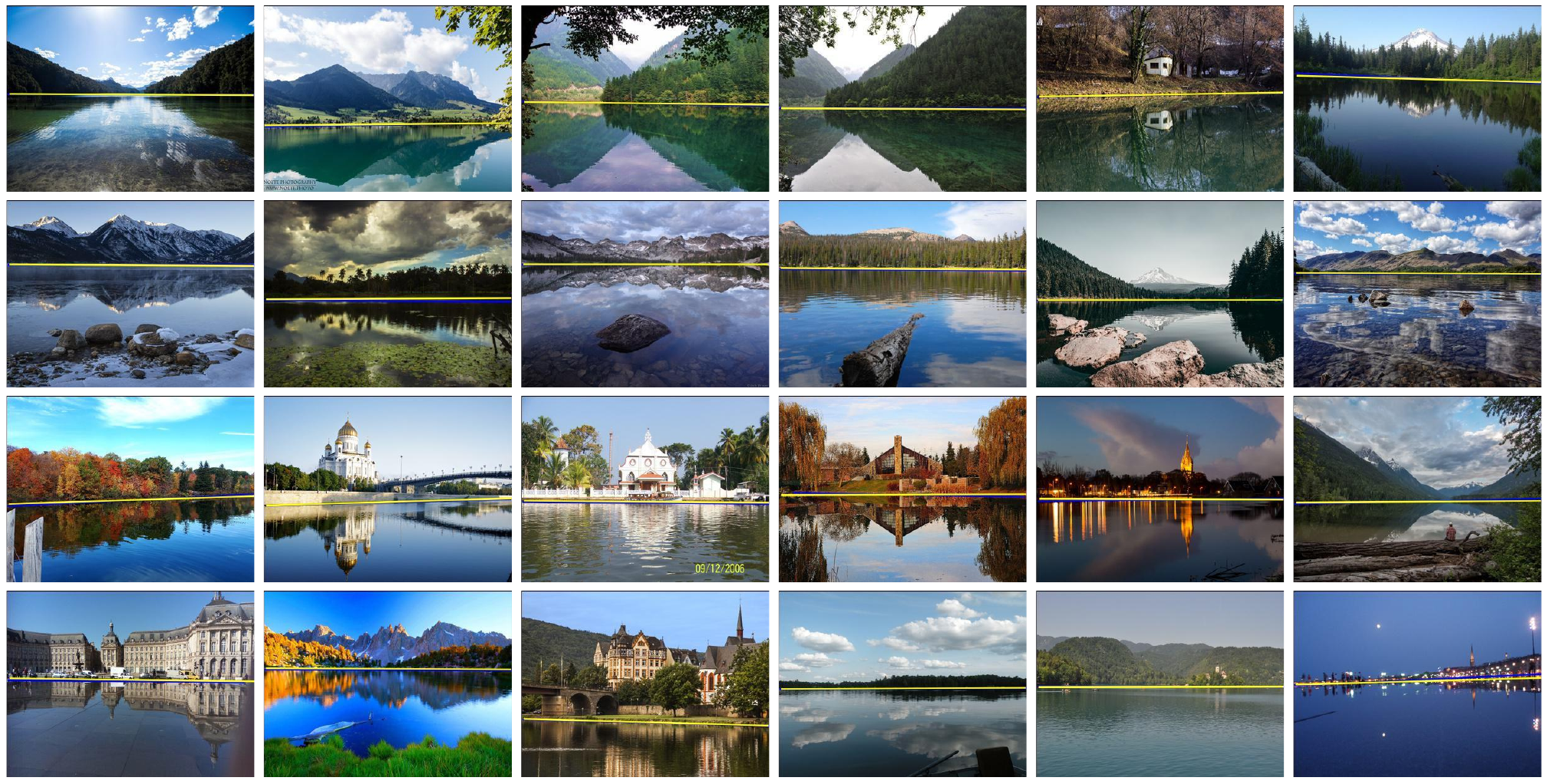}  
  \caption{Visualization of the results derived by applying the pre-trained SAWRD-Net model to 24 images that we collected.}
  \label{fig:Cross-datasetExample}
\end{figure}

To examine the generalization ability of the proposed SAWRD-Net, we collected 153 water reflection scene images from multiple online platforms, covering diverse scenarios under various illumination and weather conditions. These images were annotated using the same protocol as that used for the WRSD \cite{dong2024wrd}. The model pre-trained on the WRSD was applied to these images. A TP Rate value of 0.843 was derived, which was slightly lower than the value of 0.890 obtained on the WRSD, providing preliminary evidence of its generalization ability. Fig.~\ref{fig:Cross-datasetExample} presents the detection results of 24 newly-collected water reflectance scene images produced by the pre-trained model.

\section{Statistical Significance Tests of Performance Gains}
We repeated the three-fold cross-validation experiment seven runs by shuffling the images. Regarding each run, SAWRD-Net and WRD-Net \cite{dong2024wrd} were individually evaluated. As a result, we obtained two sets of TR Rate values: $[ 0.871, 0.886, 0.907, 0.869, \\ 0.893, 0.903, 0.871, 0.885, 0.903, 0.868, 0.892, 0.895,  0.871, 0.882, 0.895, 0.882, 0.893, \\ 0.897, 0.864, 0.893, 0.895 ]$ and $[0.797, 0.818, 0.829, 0.823, 0.815, 0.818, 0.797, 0.837, \\ 0.822, 0.780, 0.787, 0.818, 0.805, 0.822, 0.839, 0.825, 0.816, 0.801, 0.832, 0.835, 0.829]$ using SAWRD-Net and WRD-Net, respectively. On average, SAWRD-Net outperformed WRD-Net by approximately 7.00 percentage points, with a 95\% confidence interval of $[+6.19\%, +7.75\%]$. A paired $t$-test and a Wilcoxon signed-rank test \cite{wilcoxon1945individual} were performed between the two sets of data separately to investigate the statistical significance of the difference between them. In particular, the paired $t$-test yielded $t(20)=17.64$, with a corresponding $p$-value much smaller than 0.0001 ($p=1.26\times10^{-13}$), while the Wilcoxon signed-rank test also indicated a highly significant difference ($p=1.91\times10^{-6}$). The consistency between the results of the parametric and non-parametric tests demonstrates that the performance advantage of SAWRD-Net over that of WRD-Net \cite{dong2024wrd} is statistically significant. It is suggested that this advantage does not result from random fluctuations in data partitioning and remains stable across different cross-validation partitions.

\section{Training Stability of \(E(2)\)-Equivariant Convolutions}

All experimental results reported above were obtained using a fixed random seed of 42. To assess the training stability of conventional convolutions and equivariant convolutions, we randomly selected 10 different seeds and repeated the experiment. As shown in Table~\ref{tab:stability}, the equivariant convolutions achieve a standard deviation of 0.0047, which is only 41\% of that produced by conventional convolutions (0.0115), and the range is reduced by 67\%. These results demonstrate that the equivariant convolutions exhibit stronger robustness to random initialization and superior training stability compared to conventional convolutions.

\begin{table}
\renewcommand{\arraystretch}{1.5}
\begin{center}
\normalsize
\setlength{\tabcolsep}{8pt}
\caption{A comparison between conventional convolutions and equivariant convolutions in training stability across 10 random seeds.}
\label{tab:stability}
\begin{tabular}{c|c|c|c|c}
\hline
Convolutions & Mean $\uparrow$ & Std. $\downarrow$ & Var. $\downarrow$ & Range $\downarrow$ \\
\hline
Conventional & 0.7874 & 0.0115 & $1.32\times10^{-4}$ & 0.043 \\
\hline
Equivariant & \textbf{0.8861} & \textbf{0.0047} & $\mathbf{2.21\times10^{-5}}$ & \textbf{0.014} \\
\hline
\end{tabular}
\end{center}
\end{table}

\section{Ablation Studies}
\subsection{Group Number in the MSRE Block}
In the MSRE block, features produced by the \(1{\times}1\) convolution are partitioned into two subspaces: one reused for feature preservation and the other processed by a \(3{\times}3\) convolution. To validate the effectiveness of this grouping strategy, we conduct an ablation comparing the grouped architecture with its non-grouped counterpart (ReResNet), holding all other configurations identical except for the grouping operation. As shown in Table~\ref{tab:gnsrb}, the non-grouped ReResNet baseline achieves a TP Rate~\cite{nagar2019reflection} of 0.879, whereas the grouped architecture reaches 0.890, yielding an absolute gain of +0.011. These results indicate that the grouping operation contributes positively to model performance, supporting its inclusion in the complete architecture.


\begin{table}
\renewcommand{\arraystretch}{1.0}
\begin{center}
\normalsize
\setlength{\tabcolsep}{12pt}
\caption{TP Rate with different group number $s$ in the MSRE block for training.}
\label{tab:gnsrb}
\begin{tabular}{c|c}
\hline
Group Number & TP Rate \cite{nagar2019reflection} \\
\hline
$s$ = 1  & 0.879 \\
\hline
$s$ = 2& \textbf{0.890} \\
\hline
\end{tabular}
\end{center}
\end{table}

\subsection{Placement of Symmetric Attention in the Backbone}

We conducted an ablation on the placement of SA in the backbone, where \(c1\) denotes inserting SA after the first MSREG block, \(c2\) after the second, \(c3\) after the third, and \(c4\) after the fourth. The experimental results are reported in Table~\ref{tab:dpa}. Our findings show that placing SA after the first three backbone layers adversely affects overall training, whereas placing SA after the fourth layer yields a positive effect on model performance.

This behavior is consistent with the role of SA in dynamically reweighting feature maps. The first three backbone layers primarily extract low-level edge and texture features; emphasizing local regions too early can introduce noise and impair generalization. By contrast, the fourth layer produces higher-level semantic features, where SA more effectively enhances the model’s focus on critical semantic information, thereby improving detection performance.

\begin{table}
\renewcommand{\arraystretch}{1.0}
\begin{center}
\normalsize
\setlength{\tabcolsep}{12pt}
\caption{The effect of different positions of attention on TP Rate.}
\label{tab:dpa}
\begin{tabular}{ c | c }
\hline
Position & TP Rate \cite{nagar2019reflection}\\
\hline
c1 & 0.872 \\
\hline 
c2 & 0.861 \\ 
\hline 
c3 & 0.872 \\
\hline
c4 & \textbf{0.890} \\
\hline 
\end{tabular}
\end{center}
\end{table}

\subsection{Effect of Different Matrix-Decomposition Algorithms}


We evaluate three matrix-decomposition algorithms within the MD decoder under identical experimental settings, varying only the algorithm itself. The candidates are Nonnegative Matrix Factorization (NMF)~\cite{lee1999learning}, Vector Quantization (VQ)~\cite{gray1998quantization}, and Concept Decomposition (CD)~\cite{dhillon2001concept}. As shown in Table~\ref{tab:dmda}, the CD-based model attains a TP Rate of 0.885, the VQ-based model achieves 0.887, and the NMF-based model reaches 0.890. These results indicate that NMF is the most suitable among the tested algorithms for the water reflection detection task, with the NMF variant yielding the best overall performance.

\begin{table}[!htb]
\renewcommand{\arraystretch}{1.0}
\begin{center}
\normalsize
\setlength{\tabcolsep}{12pt}
\caption{Effect of different matrix decomposition algorithms in the decoder on TP Rate during training.}
\label{tab:dmda}
\begin{tabular}{ c | c }
\hline
MD Algorithms & TP Rate \cite{nagar2019reflection}\\
\hline
CD \cite{dhillon2001concept} & 0.885 \\
\hline 
VQ \cite{gray1998quantization} & 0.887\\ 
\hline 
NMF \cite{lee1999learning} & \textbf{0.890}\\
\hline 
\end{tabular}
\end{center}
\end{table}

\bibliographystyle{elsarticle-num}
\bibliography{sawrd-net}